%% file: main.tex
\documentclass[journal]{IEEEtran}
\IEEEoverridecommandlockouts

\usepackage{amssymb}
\usepackage{pifont}
\usepackage{amsmath}
\usepackage{silence}
\usepackage[hidelinks]{hyperref}
\usepackage{comment}
\usepackage[acronym]{glossaries-extra}
\setabbreviationstyle[acronym]{long-short}
\glssetcategoryattribute{acronym}{nohyperfirst}{true}

\usepackage{booktabs}
\usepackage[usenames,dvipsnames]{color}
\newif\ifshowrevisions
\showrevisionsfalse   

\newcommand{\rev}[1]{%
  \ifshowrevisions
    \textcolor{blue}{#1}%
  \else
    #1%
  \fi
}
\usepackage{array}
\usepackage{cite}
\usepackage{multirow}
\usepackage{adjustbox}
\usepackage{longtable}
\usepackage{float}
\usepackage{threeparttable}
\usepackage{subcaption}
\usepackage[capitalize]{cleveref}

\input{symbols}

\begin{document}



\title{Reproducibility in the Control of Autonomous Mobility-on-Demand Systems}

\author{Xinling Li$^{1}$, Meshal Alharbi$^{1}$, Daniele Gammelli$^{2}$, James Harrison$^3$, Filipe Rodrigues$^4$, Maximilian Schiffer$^5$, Marco Pavone$^2$, Emilio Frazzoli$^6$, Jinhua Zhao$^{7}$, Gioele Zardini$^{1}$ 

        \thanks{$^1$Laboratory for Information and Decision Systems, Massachusetts Institute of Technology, Cambridge, MA 02139 USA,
        {\tt \{xinli831,meshal,gzardini\}@mit.edu}}
        \thanks{
        $^2$Department of Aeronautics and Astronautics, Stanford University, CA 94305 USA,
        {\tt \{gammelli,pavone\}@stanford.edu}}
        \thanks{
        $^3$Google DeepMind, San Francisco, CA, USA, 
        {\tt jamesharrison@google.com}}
        \thanks{
        $^4$Technical University of Denmark, Kongens Lyngby, Denmark,
        {\tt rodr@dtu.dk}}
        \thanks{
        $^5$School of Management, Technical University of Munich, Germany,
        {\tt schiffer@tum.de}}
        \thanks{
        $^6$Institute for Dynamic Systems and Control, ETH Zurich, Switzerland,
        {\tt frazzoli@ethz.ch}}
        \thanks{
        $^7$Department of Urban Studies and Planning, Massachusetts Institute of Technology, Cambridge, MA 02139 USA,
        {\tt jinhua@mit.edu}}
        \thanks{This work is supported by the US Department of Energy/Office of Energy Efficiency and Renewable Energy, Grant Agreement DE-EE0011186, the Sidara Urban Seed Grant Program at the Norman B. Leventhal Center for Advanced Urbanism, Massachusetts Institute of Technology, and by Prof. Zardini's grant from the MIT Amazon Science Hub, hosted in the Schwarzman College of Computing.}
}
\markboth{}%
{Shell \MakeLowercase{\textit{et al.}}: A Sample Article Using IEEEtran.cls for IEEE Journals}

\maketitle
\begin{abstract}
\gls{abk:amod} systems, powered by advances in robotics, control, and \gls{abk:ml}, offer a promising paradigm for future urban transportation.
\gls{abk:amod} offers fast and personalized travel services by leveraging centralized control of autonomous vehicle fleets to optimize operations and enhance service performance.
However, the rapid growth of this field has outpaced the development of standardized practices for evaluating and reporting results, leading to significant challenges in reproducibility.
As \gls{abk:amod} control algorithms become increasingly complex and data-driven, a lack of transparency in modeling assumptions, experimental setups, and algorithmic implementation hinders scientific progress and undermines confidence in the results.

This paper presents a systematic study of reproducibility in \gls{abk:amod} research.
We identify key components across the research pipeline, spanning system modeling, control problems, simulation design, algorithm specification, and evaluation, and analyze common sources of irreproducibility.
We survey prevalent practices in the literature, highlight gaps, and propose a structured framework to assess and improve reproducibility.

While focused on \gls{abk:amod}, the principles and practices we advocate generalize to a broader class of cyber-physical systems that rely on networked autonomy and data-driven control.
This work aims to lay the foundation for a more transparent and reproducible research culture in the design and deployment of intelligent mobility systems.
\end{abstract}


\input{chapters/introduction}

\input{chapters/model}

\input{chapters/operation}

\input{chapters/algorithm}

\input{chapters/assumption}

\input{chapters/evaluation}

\input{chapters/experiment}

\input{chapters/interaction}

\input{chapters/conclusion}

\bibliographystyle{IEEEtran}

{\footnotesize
\bibliography{cas-refs}}
\end{document}

%% file: symbols.tex
\newacronym{abk:amod}{AMoD}{Autonomous Mobility-on-Demand}
\newacronym{abk:mod}{MoD}{Mobility-on-Demand}
\newacronym{abk:av}{AV}{Autonomous Vehicle}
\newacronym{abk:gui}{GUI}{Graphical User Interface}
\newacronym{abk:co}{CO}{Combinatorial Optimization}
\newacronym{abk:od}{OD}{Origin-Destination}
\newacronym{abk:rl}{RL}{Reinforcement Learning}
\newacronym{abk:ev}{EV}{Electric Vehicle}
\newacronym{abk:v2g}{V2G}{Vehicle-to-Grid}
\newacronym{abk:mpc}{MPC}{Model Predictive Control}
\newacronym{abk:mdp}{MDP}{Markov Decision Process}
\newacronym{abk:bpr}{BPR}{Bureau of Public Roads}
\newacronym{abk:mfd}{MFD}{Macroscopic Fundamental Diagram}
\newacronym{abk:ctm}{CTM}{Cell Transmission Model}
\newacronym{abk:ltm}{LTM}{Link Transmission Model}
\newacronym{abk:bpm}{BMP}{Bipartite Matching Problem}
\newacronym{abk:marl}{MARL}{Multi-Agent Reinforcement Learning}
\newacronym{abk:ml}{ML}{Machine Learning}

\newcommand{\st}{\text{s.t. }}

%% file: chapters/introduction.tex
\section{Introduction}
The rapid urbanization over the past decades has led to a significant rise in private automobile use.
While private vehicles offer convenient, door-to-door service at any time, their proliferation has resulted in widespread traffic congestion, increased emissions, and a substantial loss of productive hours~\cite{schrank20112011}. 
In response, \gls{abk:mod} services have emerged as a compelling alternative.
By dynamically allocating vehicles based on real-time demand, \gls{abk:mod} systems provide personalized travel options while reducing dependence on private car ownership.
As such, they retain many of the advantages of private vehicles, offering a more sustainable and scalable mobility solution.
With advances in autonomous driving technologies, \gls{abk:amod} has gained traction as a key enabler of next-generation transportation systems~\cite{Pavone2015}.
\gls{abk:amod} systems (i.e., fleets of robo-taxis performing on-demand mobility tasks) autonomously manage vehicle fleets to respond to passenger requests and satisfy transportation demand effectively.
Such systems are inherently complex: they involve high degrees of interaction and require coordination across spatial and temporal scales\cite{zardini2022analysis}.
As a result, different studies often focus on distinct operational challenges, each adopting specific assumptions regarding system dynamics and passenger behavior.
Such methodological diversity complicates efforts to ensure research reproducibility and benchmark comparability.
Nevertheless, reproducibility remains a cornerstone of rigorous scientific practice.
It increases credibility, accelerates progress, and enables meaningful comparisons across research contributions~\cite{munafo2017manifesto}. 
In disciplines such as computational science and management science, reproducibility is widely promoted via standardized evaluation procedures, common benchmarks, and the routine of sharing code and data~\cite{peng2011reproducible,leitner2017acrv,baines2024need,fivsar2024reproducibility}. 
By contrast, while the importance of reproducibility is gaining recognition in transportation research with networked agents\cite{wu2024reproducibility, riehl2025revisiting}, consistent practices and community standards are still scarce.

This paper aims to address this gap by identifying and analyzing the key components that determine the reproducibility of \gls{abk:amod} control research.
\rev{\gls{abk:amod} is \emph{among the first and most visible} real-world, \emph{city-scale} deployments of networked robot autonomy in open, human-populated environments, crystallizing the field’s shift \emph{from car to fleet}: the phenomena we analyze, matching, dispatching, and rebalancing, are precisely the robotics primitives of multi-robot task allocation, online routing/trajectory coordination, and load balancing under uncertainty.
While the focus is on \gls{abk:amod}, the principles discussed are applicable to a broad range of networked control and optimization problems, including those in aviation, maritime transport, and other complex cyber-physical systems.}

The term \textit{reproducibility} has been used to describe varying degrees of rigor in how scientific findings are communicated and validated \cite{goodman2016does, national2019reproducibility}. At its most basic level, reproducible research requires sufficient transparency and methodological clarity to enable others to recreate the results and verify the underlying logic. This minimal standard typically assumes access to the same input data and requires that the described methods and conditions are detailed enough to reproduce the original outcomes. A more rigorous standard is achieved when independent researchers, using comparable techniques and assumptions and not necessarily the same data, can arrive at similar findings and inferences. This higher level, which is sometimes referred to as inferential reproducibility or replicability, is critical not only for confirming the robustness of scientific claims but also for enabling future research to build upon and extend the original work. In this article, we strive to provide recommendations that strengthen both levels of reproducibility of \gls{abk:amod} research, though we do not draw a strict distinction between the terms.

\subsection{Holistic, System-level Reproducibility for AMoD}
\gls{abk:amod} systems operate fleets of \glspl{abk:av} to fulfill passenger requests within a transportation network.
Users request trips via mobile applications, and the system must coordinate vehicles in real time to deliver a high-quality service.
The \gls{abk:amod} control problem, therefore, centers on optimizing fleet management, with performance metrics including profitability, service rate, passenger waiting time, and environmental impact.

\begin{figure*}[tbh]
\centering
\includegraphics[width=0.9\textwidth]{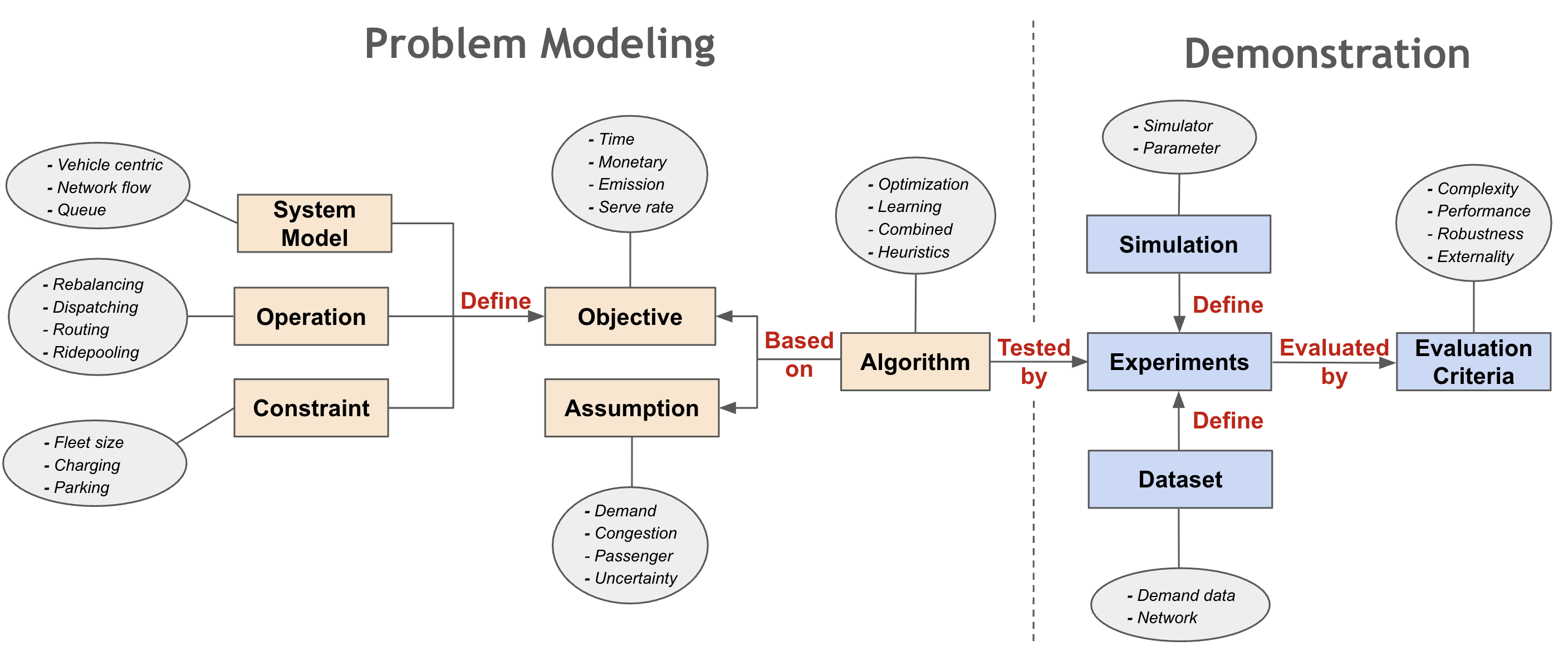}
\caption{The reproducibility scheme. The graph includes key components to improve \gls{abk:amod} control research reproducibility.}\label{fig:diagram}
\end{figure*}

\cref{fig:diagram} presents a scheme of the system-level reproducibility of \gls{abk:amod} systems, highlighting the essential components needed to improve reproducibility in this context.
Specifically, it includes not only the modeling and algorithmic aspects of specific research efforts, but also key experimental and contextual details, such as assumptions, constraints, and data sources, often under-reported.
Together, these components provide a comprehensive blueprint for communicating research transparently and facilitating reproducibility.

To define an \gls{abk:amod} control problem, three key elements must be specified: (1) the system model, (2) the target operation, and (3) the control objective.
The system model provides a tractable mathematical representation of the real-world \gls{abk:amod} network and serves as the foundation for problem formulation.
Based on such a model, researchers define the operation of interest, typically some aspect of fleet control constrained by infrastructure properties (e.g., road capacity), and select an appropriate objective function, which often reflects operational efficiency, profitability, or sustainability.
Crucially, any proposed solution is meaningful only in the precise context of its assumptions.
These include both explicit modeling assumptions (e.g., demand elasticity, congestion models, penetration rates of new technologies), and implicit design choices (e.g., simulation fidelity).
These assumptions directly affect the applicability, generalizability, and replicability of the results.
Once a solution is developed, it should be evaluated through experiments that benchmark performance, demonstrate robustness, and assess computational tractability.
Clear reporting of all steps in this pipeline is essential to support reproducible research.

While prior surveys on \gls{abk:amod} control have focused on specific operational problems or solution approaches~\cite{9924240,zardini2022analysis}, this is, to the best of our knowledge, the first perspective centered on reproducibility.
In the remainder of this work, we elaborate on the components reported in \cref{fig:diagram}, discussing current practices, identifying common pitfalls, and proposing guidelines for improvement.

\subsection{Organization of the Manuscript}
The remainder of this paper is organized as follows. 
\cref{sec:sys} presents modeling approaches for \gls{abk:amod} systems. 
\cref{sec:operation-all} discusses key operational challenges. 
\cref{sec:algo} reviews algorithmic techniques used to address these challenges. \cref{sec:assumption} outlines foundational assumptions about system dynamics and their implications for research applicability. \cref{sec:evaluation} summarizes the metrics often used for benmarking.
Furthermore, \cref{sec:experiment} details best practices for experimental design and documentation. 
\cref{sec:interaction} then explores studies on interactions between \gls{abk:amod} systems and infrastructures.
Finally,  \cref{sec:conclusion} concludes the work with a reproducibility checklist as a reference for future research. 

%% file: chapters/model.tex
\section{Mathematical Modeling AMoD Systems} \label{sec:sys}
\rev{The essence of modeling \gls{abk:amod} systems is to capture the dynamic interplay between passengers, vehicles, and their interactions within the transportation network.
The system model serves as the foundation for any resource allocation strategy by formally defining the entities involved, the resources (\glspl{abk:av}), the consumers (passengers), and the logic governing their interactions.
In the existing \gls{abk:amod} literature, most system models fall into three categories: queueing-theoretic, network flow models, and vehicle-centric (agent-based) models, each built on distinct assumptions and tailored to highlight different operational characteristics of \gls{abk:amod} systems.
This section introduces these three modeling approaches, focusing on their assumptions, advantages, and limitations, and explains why understanding these trade-offs is critical both for selecting an appropriate framework and for interpreting the applicability and reproducibility of derived results. 
We close by encouraging future work to explicitly specify the chosen system model, including its assumptions and structural limitations.} 
%

\subsection{Queueing-theoretic models}
\rev{Queueing theory provides a mathematical framework for analyzing waiting lines and service systems.
In \gls{abk:amod}, queueing-theoretic models typically represent trip requests as stochastic arrival processes with vehicles picking up passengers.
Network locations are modeled as queues, and system dynamics are described by the flow of vehicles and passengers between them~\cite{zhang2018analysis,zhang2016control}.}
\subsubsection{Assumptions} 
\label{sec:assumptions-queue}
\rev{These models represent passengers and vehicles as \emph{discrete} entities. Customers arrive according to an exogenous stochastic process, reaching each origin station at a \emph{fixed} rate and selecting destinations according to predefined probabilities~\cite{iglesias2019bcmp,braverman2019empty}.
To cast \gls{abk:amod} within the queueing framework, additional assumptions are imposed on demand, travel times, and passenger behavior: demand is commonly time-invariant and Poisson, travel times between stations are exponentially distributed, and passengers immediately leave the system if not matched with a vehicle~\cite{zhang2016control,zhang2018analysis,iglesias2019bcmp}. 
These assumptions enable analytical treatment while preserving accuracy: \cite{larson1981urban} shows that reasonable deviations from them have negligible impact on performance.}



\subsubsection{Benefits} 
\rev{Queueing-theoretic models preserve the discrete nature of passenger arrivals and allow the direct use of queueing theory to compute key performance metrics, such as node throughput~\cite{shortle2018fundamentals}, which corresponds to vehicle movement and availability in \gls{abk:amod} systems. They explicitly capture system randomness and guarantee performance in expectation, and the resulting policies maintain system equilibrium, preventing queues and waiting times from growing unbounded~\cite{iglesias2019bcmp,zhang2016control,zhang2018analysis}.}


\subsubsection{Limitations}
\rev{As discussed in \cref{sec:assumptions-queue}, queueing-theoretic models rely on strong assumptions regarding passenger arrival rates and travel times, which are necessary for analytical tractability. 
Arrival rates are often assumed time-invariant to enable steady-state analysis.
However, these simplifications may fail to capture real-world transportation dynamics, particularly in long-term planning scenarios where demand is non-stationary and influenced by external factors.}
%
%

\subsection{Network flow models}
\rev{Network flow models are among the most widely adopted approaches in \gls{abk:amod} research, and represent the transportation system as a directed graph, where nodes correspond to physical or logical locations and arcs to possible vehicle or passenger movements, with system dynamics captured by flows of vehicles and passengers across the network.}

\subsubsection{Assumptions}
\rev{When paired with fluid dynamics~\cite{pavone2012robotic}, network flow models treat vehicle and passenger flows as continuous variables. 
Although this abstraction departs from the discrete nature of individual trips and vehicles, it enables substantial computational simplifications: problems originally formulated as integer programs can often be relaxed to linear programs, supporting scalable optimization even on large networks. 
For large transportation systems, recent studies show that the difference between microscopic simulation and continuous network flow models is marginal at the system level~\cite{molter2024public,lienkamp2024column}, further motivating their use.}

\subsubsection{Benefits} 
\rev{These models align naturally with foundational concepts in transportation science, such as demand modeling via \gls{abk:od} pairs and network equilibrium analysis. 
Their graph-based structure provides an intuitive, modular framework for capturing fleet movements, demand fulfillment, and operational constraints. With a sufficiently fine node discretization, network flow models can approximate free-floating \gls{abk:amod} systems with high fidelity and minimal discretization error~\cite{spieser2016shared}.}
%
\subsubsection{Limitations} 
\rev{Despite their widespread use and conceptual clarity, network flow models face challenges when translating abstract flows into actionable control policies. 
Control decisions are typically expressed as aggregate flows between nodes rather than explicit vehicle trajectories, so an additional path-reconstruction step is required to derive path-level instructions for \glspl{abk:av}. 
While often bypassed in simulation studies, this step is critical in real-world implementations, where routing feasibility, vehicle-level state, and road network constraints (e.g., capacity) must be respected~\cite{wollenstein2021routing}.}


\subsection{Vehicle-centric models}
Unlike network flow or queueing-theoretic models that abstract away vehicle identities in favor of aggregate flows, this approach explicitly tracks the state and control actions of every vehicle in the system.
\subsubsection{Assumptions} 
\rev{Vehicle-centric models assume that system dynamics are described and controlled at the individual-vehicle level: each vehicle has a state (location, occupancy, battery level, assigned task such as serving a customer or rebalancing)~\cite{enders2023hybrid,alonso2017demand,boewing2020vehicle}, and control actions (dispatching, routing, rebalancing) are chosen per vehicle, often based on local or partially observed information.}

%
\subsubsection{Benefits} 
\rev{This paradigm yields direct, interpretable, and immediately executable per-vehicle decisions, simplifying deployment in real-world fleet management systems; unlike network flow models, it does not require a path reconstruction step to translate aggregate flows into individual actions, and it naturally accommodates heterogeneous fleets and environments, including vehicle-specific capacities, preferences, and constraints.}


\subsubsection{Limitations} 
\rev{Under centralized control, scalability is the main limitation: decision variables grow linearly with fleet size, whereas queueing and network flow models typically scale with the number of \gls{abk:od} pairs, leading to high computational burden and reduced tractability for large systems; distributed algorithms can mitigate this but may degrade decision quality and overall fleet performance.}


\subsection{Reproducibility considerations}
\rev{Modeling choices critically affect the reproducibility and comparability of \gls{abk:amod} control research, since each paradigm entails implicit assumptions shaping problem formulation, solution methods, and result interpretation. When these assumptions are not explicit, studies are hard to replicate or compare—for example, queueing models may hide how time-varying demand is handled, network flow models may omit path reconstruction details, and vehicle-centric models may underspecify vehicle states and constraints. To support reproducibility, researchers should clearly document the chosen modeling framework, its structural assumptions and abstractions, and any simplifications made for tractability.}


%% file: chapters/operation.tex
\section{Operating AMoD Systems} \label{sec:operation-all}
\label{sec:operating-amod}
\rev{Deploying \gls{abk:amod} systems requires real-time operational policies that ensure efficient service by \emph{maximizing} performance, typically via dispatching vehicles for pickup and delivery and proactively repositioning them to anticipated high-demand areas, subject to system-specific constraints. 
This section defines the key operational tasks and constraints governing \gls{abk:amod} systems, aiming to establish a clear reference rather than provide a comprehensive review of existing solutions.}
\subsection{Operational tasks}
\label{sec:operation}
\begin{figure}[tb]
\begin{center}
\begin{subfigure}[tb]{0.45\linewidth}
\includegraphics[width=\linewidth]{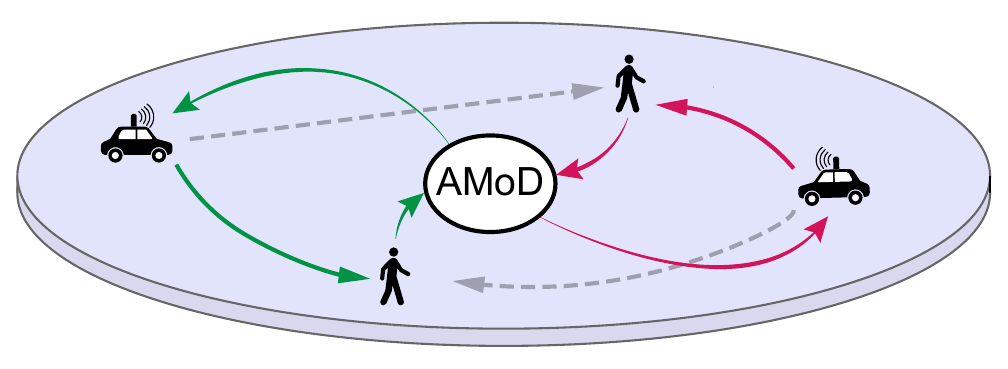}
\caption{Matching. \label{fig:dispatching}}
\end{subfigure}
\begin{subfigure}[tb]{0.45\linewidth}
\includegraphics[width=\linewidth]{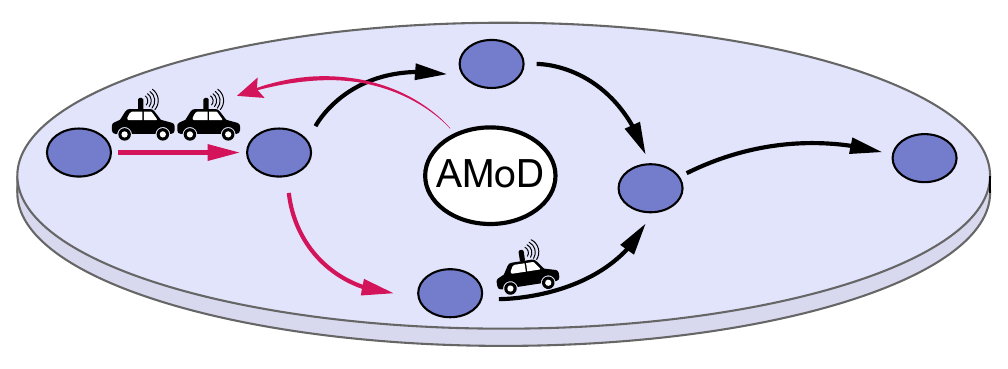}
\caption{Rebalancing/Relocation. \label{fig:rebalancing}}
\end{subfigure}
\\
\begin{subfigure}[tb]{0.45\linewidth}
\includegraphics[width=\linewidth]{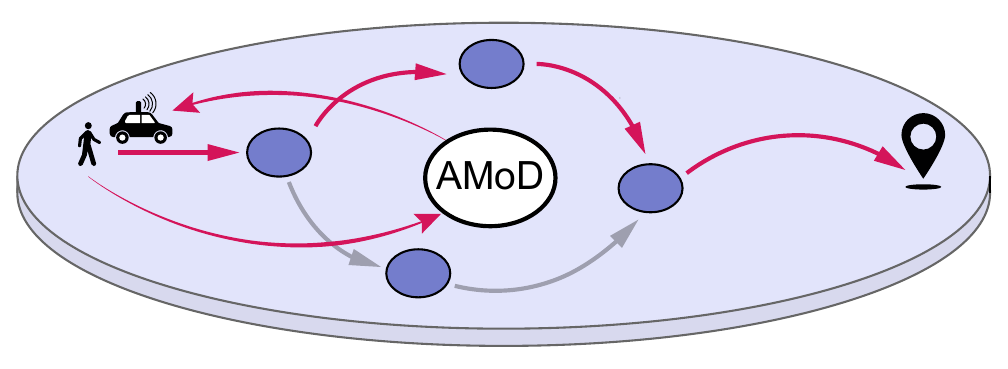}
\caption{Routing. \label{fig:routing}}
\end{subfigure}
\begin{subfigure}[tb]{0.45\linewidth}
\includegraphics[width=\linewidth]{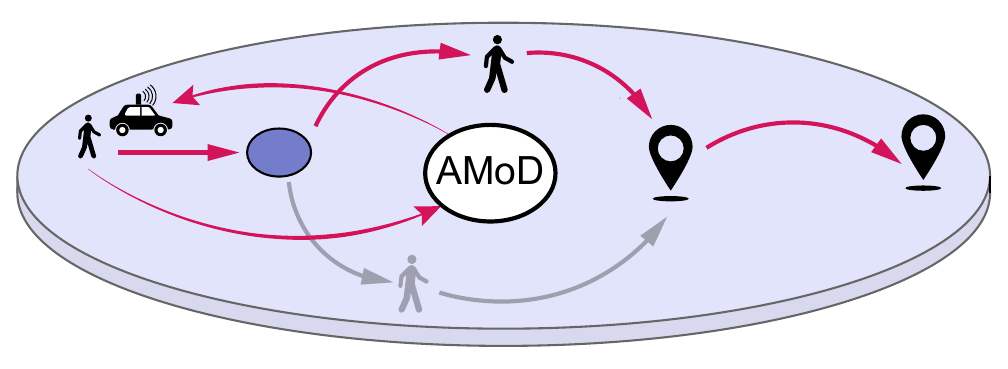}
\caption{Ridepooling. \label{fig:ridepooling}}
\end{subfigure}
\end{center}
\caption{Operational problems in \gls{abk:amod} control~\cite{zardini2022analysis}.}
\label{fig:op_policies}
\end{figure}

Operational tasks include dispatching, rebalancing, routing, and ridepooling, graphically reported in \cref{fig:op_policies}.
For an in-depth review of operational strategies, the reader can refer to~\cite{zardini2022analysis}.
\subsubsection{Matching}
\rev{Given the set of passengers and vehicles available, matching focuses on the assignment of vehicles to the passengers (\cref{fig:dispatching}). 
In the one-to-one passenger-to-vehicle assignment setting, the matching problem can be formulated as a \gls{abk:bpm}~\cite{enders2023hybrid,woywood2024multi,hu2020artificial}, where the arc costs are related to the control objectives.}
\subsubsection{Rebalancing}
\rev{Vehicle rebalancing, or relocation, is the redistribution of available vehicles to areas with supply–demand imbalances (\cref{fig:rebalancing}). 
By optimizing fleet distribution, it improves service rates and utilization, and can reduce average wait times and delays~\cite{wallar2018vehicle}, with particularly large benefits when vehicle supply is scarce~\cite{li2025learning}.}

\subsubsection{Vehicle routing}
\rev{The vehicle routing problem focuses on scheduling vehicle visits to stops given a set of tasks (\cref{fig:routing}); see~\cite{schiffer2019vehicle} for a comprehensive review. 
In \gls{abk:amod} systems, routing arises at two levels. 
At the road network level, given an \gls{abk:od} pair, the goal is to select routes for vehicles, where algorithmic solutions can go beyond naive shortest paths and exploit fleet-level routing to balance flows across alternative paths~\cite{jalota2023online} or stagger them temporally~\cite{coppola2024staggered} to avoid congestion bottlenecks~\cite{lazar2021learning}. 
At the vehicle operation level, routing determines the sequence in which vehicles visit stations under dispatch and/or rebalancing decisions; in the \gls{abk:amod} literature, the term ``routing'' usually refers to the latter.}

\subsubsection{Charging scheduling}
\rev{Growing sustainability concerns are accelerating the adoption of \gls{abk:ev}, with projections indicating a 12-fold increase by 2035~\cite{iea2024globalEV}, making electrification a key component of future mobility systems. 
Compared to conventional \gls{abk:amod}, electric \gls{abk:amod} introduces additional charging-related operational tasks that must be incorporated into fleet coordination. 
A central research direction is active charging schedule design that models the system-level interaction between electric \gls{abk:amod} fleets and power grids, typically by bounding charging requests according to available power supply~\cite{boewing2020vehicle} or integrating \gls{abk:v2g} capabilities, where \glspl{abk:ev} can both draw from and feed energy back into the grid~\cite{iacobucci2018modeling,iacobucci2019optimization}. 
Beyond jointly modeling charge scheduling and power network impacts, fleet-wide charge scheduling driven by price signals is emerging as a promising avenue to increase system efficiency~\cite{klein2023electric}.}

\subsubsection{Ridepooling}
\rev{In \gls{abk:amod}, ridepooling allows multiple passengers to share a vehicle (\cref{fig:ridepooling}), improving service efficiency by reducing required fleet size, energy use, and congestion~\cite{bei2018algorithms,narayanan2020shared}. 
It is typically categorized by vehicle capacity:
a) Low capacity ($\leq 3$): vehicles accommodate up to 3 passengers~\cite{jindal2018optimizing,cap2018multi,enders2023hybrid,tsao2019model,yu2019integrated}; 
b) High capacity ($\geq 4$): vehicles accommodate 4 or more passengers~\cite{tong2018unified,gueriau2018samod,al2019deeppool,zheng2018order,alonso2017predictive,alonso2017demand,fielbaum2021demand,kim2024learning,shah2020neural,fagnant2018dynamic,simonetto2019real,gueriau2020shared,wallar2018vehicle,wen2017rebalancing}.
This distinction is crucial, as capacity directly affects the scalability of solution approaches: in low-capacity settings, control strategies often resemble those of single-passenger systems and remain tractable~\cite{alonso2017demand, li25speedup}, whereas high-capacity ridepooling introduces substantial computational challenges, typically addressed via heuristics~\cite{li2021optimal,alonso2017demand} or simplifications such as first-come-first-served allocation~\cite{fagnant2018dynamic} or limiting assignments to one passenger per time step~\cite{kim2024learning,jindal2018optimizing}. 
A related design choice is between single-request and batch assignment: single-request schemes assign at most one new passenger per vehicle at each decision epoch~\cite{kim2024learning,enders2023hybrid,simonetto2019real}, simplifying the problem by ignoring temporal coupling across requests, whereas batch schemes match multiple passengers and vehicles simultaneously~\cite{shah2020neural,alonso2017demand}, capturing such dependencies at the cost of higher computational complexity.}

\subsubsection{Reproducibility considerations}
\rev{Inconsistent terminology in the \gls{abk:amod} control literature creates confusion and obscures which operational problems are actually being addressed, hindering comparability and reproducibility. We therefore propose the following guidelines.}
\paragraph{Matching vs. dispatching vs. rebalancing}
\rev{The term \emph{dispatching} is often (ambiguously) used to mean \emph{matching}, and at times even interchangeably with \emph{rebalancing}~\cite{chen2017hierarchical}, despite referring to distinct operations. 
We recommend using \emph{matching} exclusively for assigning vehicles to passenger requests, and \emph{rebalancing} for relocating idle vehicles to areas of anticipated demand, while reserving \emph{dispatching} as a generic term only when explicitly defined.}

\paragraph{Routing}
\rev{In \gls{abk:amod} systems, routing decisions may result from dispatching, rebalancing, or both. 
Using \emph{routing} as a catch-all for the operational problem can obscure underlying decisions. 
One should explicitly state which operations (e.g., dispatch, rebalancing) are optimized when referring to routing.}
\paragraph{Shared vs. ridepooling}
\rev{\emph{Shared} \gls{abk:amod} is sometimes used for fleets shared among users but serving only one passenger at a time~\cite{dandl2019evaluating,kang2021maximum,levin2017congestion}. 
To avoid ambiguity, we recommend reserving \emph{ridepooling} for shared trips where multiple passengers travel simultaneously in the same vehicle.}
\subsection{Operational constraints}
Solving the control tasks in \cref{sec:operation} requires accounting for operational constraints imposed by system-specific characteristics, fundamental in real-world deployments.

\subsubsection{Fleet size}
\rev{Most studies control a given fleet, treating its size as a constraint~\cite{kang2021maximum,yang2020planning,zardini2022co}. 
How this enters the model depends on the paradigm: in vehicle-centric models it is implicit, as decisions are made per vehicle; in queueing models it is inherent in the closed network structure; and in network flow models it is enforced via an explicit cap on available vehicles, typically computed as the sum of vehicles in transit~\cite{yang2020planning,kang2021maximum} or derived from initial conditions~\cite{pavone2012robotic}. 
Other works treat fleet size as a design variable, optimizing it based on system performance metrics~\cite{vazifeh2018addressing,ma2017designing,volkov2012markov,li2019efficient,fagnant2018dynamic,spieser2016vehicle}.}
\subsubsection{Charging}
\rev{Beyond proactive charging as an operational task, many studies model charging as a constraint that limits fleet coordination due to the restricted range of \glspl{abk:ev}, determined by battery capacity and energy consumption during travel~\cite{hu2020artificial,belakaria2019multi,guo2020vehicle}. 
In addition to fixed charging layouts, the design and placement of charging infrastructure remain critical research topics~\cite{chen2013electric,jia2012optimal,zhang2016pev}, with joint optimization of station locations and fleet operations enabling more efficient system-wide coordination~\cite{luke2021joint}. 
Similar considerations apply to general maintenance activities (e.g., cleaning, calibration)~\cite{zardini2022analysis}.}

\subsubsection{Parking}
\rev{Transportation activities typically begin and end with parking, making parking space a critical resource, especially in constrained urban areas~\cite{inci2015review}. 
Parking can be modeled via hard capacity limits on the number of vehicles per node~\cite{chu2021joint,levin2017congestion,nair2011fleet} or through additional idling costs that discourage long dwell times and implicitly capture parking scarcity~\cite{becker2020impact,sieber2020improved,ma2017designing}. Nonetheless, most \gls{abk:amod} control studies assume unlimited parking, so pickup and idling are unconstrained by capacity, overlooking real-world challenges.}
\subsubsection{Reproducibility considerations}
\rev{Operational constraints are central to the applicability of \gls{abk:amod} control methods but are often under-specified or inconsistently communicated. They encode both the scenario of interest and the reality gap of a proposed approach, and should therefore be clearly listed and tied to their mathematical encoding. 
While charging constraints are usually explicit, signaling a focus on \glspl{abk:ev}, constraints on fleet size and parking are often less transparent and left to be inferred from the solution method. 
Explicitly stating all such constraints, even when they are implicit in the modeling or optimization framework, greatly improves clarity, reproducibility, and the ability to benchmark results.}

%% file: chapters/algorithm.tex
\section{Algorithms for AMoD Control} \label{sec:algo}
\rev{Designing effective control policies for \gls{abk:amod} fleets hinges on algorithms that address the operational challenges outlined in \cref{sec:operation}, an area that has recently attracted interest from robotics, control, transportation, operations research, and machine learning. We group existing approaches into four categories: mathematical programming, learning-based methods, hybrid approaches, and customized heuristics, each with distinct modeling assumptions, solution strategies, and trade-offs between optimality, scalability, and interpretability. 
This section outlines the core ideas behind these algorithm classes and surveys their use in \gls{abk:amod} control problems, emphasizing that algorithm design, formulation, and implementation details must be clearly reported to support reproducibility.}
\subsection{Mathematical programming}
\rev{Mathematical programming is a core tool in operations research and decision making in complex systems. 
It formulates a system model with an objective function and constraints capturing system dynamics and operational requirements; time is typically discretized into decision intervals, and control actions are obtained by solving an optimization problem at each interval~\cite{bellman1966dynamic}. In \gls{abk:amod} control, objectives usually target service quality and profitability (see \cref{sec:evaluation}), while constraints depend on the underlying network model and operations. 
In particular, rebalancing is commonly modeled via network flow~\cite{pavone2012robotic,rossi2018routing,volkov2012markov,salazar2019congestion} or queue-theoretic formulations~\cite{iglesias2019bcmp,zhang2018analysis,zhang2016control,braverman2019empty,belakaria2019multi}.}

\subsubsection{Rebalancing with network flow and queue-theoretical models}
\rev{Consider a system with~$N$ stations and discretized decision windows~$t \in \{1,\dots,T\}$, where rebalancing decisions are made over \gls{abk:od} pairs. Let~$p_{ij}$ and~$c_{ij}$ denote the profit from a passenger-carrying trip and the cost of rebalancing a vehicle from station~$i$ to~$j$, respectively; $s_i^t$ the number of vehicles at station~$i$ at time~$t$;~$x_{ij}^t$ the number of empty vehicles rebalanced from~$i$ to~$j$ at time~$t$;~$y_{ij}^t$ the number of customer-carrying vehicles from~$i$ to~$j$ at time~$t$; $d_{ij}^t$ the customer demand; and~$t_{ij}$ the travel time. 
The network-flow based optimal rebalancing problem is represented as:}
%
\begin{subequations}
    \begin{align}
    \max & \quad \sum_{i,j,t}  p_{ij}y_{ij}^t - \sum_{i,j,t}  c_{ij}x_{ij}^t \label{eq:reb_obj}\\
    \st& \quad y_{ij}^{t} \leq d_{ij}^{t} \quad \forall i \in N,\forall t \in T, \label{eq:reb_st1}\\
    & \quad     s_i^{t+1} = s_i^t + \sum_{j=1}^{n} x_{ji}^{t-t_{ji}} + \sum_{j=1}^{n} y_{ji}^{t-t_{ji}} - \sum_{j=1}^{n} x_{ij}^t \label{eq:reb_st2}\\ & - \sum_{j=1}^{n} y_{ij}^t \quad \forall i \in N, \forall t \in T, \nonumber \\    
    & \quad x_{ij}^t,y_{ij}^t,s_{i}^{t} \in \mathbb{N} \quad \forall i,j \in N, \forall t \in T,
    \end{align}
\end{subequations}
\rev{where \cref{eq:reb_st1} enforces that served demand does not exceed arrivals, \cref{eq:reb_st2} enforces flow conservation, and \cref{eq:reb_obj} maximizes service profit.}

\rev{Queueing-theoretical models impose constraints tailored to network-level goals such as stability~\cite{belakaria2019multi} and balanced resource allocation~\cite{iglesias2019bcmp,zhang2016control,zhang2018analysis}. 
Using the same network and demand setup as in network flow models, a Jackson network-based formulation~\cite{zhang2016control} minimizes rebalancing time while equalizing service rates across stations:}
\begin{subequations}
\begin{align}
    \min & \quad \sum_{i,j} t_{ij}\alpha_{ij}\psi_i\label{eq:queue_obj}\\
    \st& \quad \gamma_i = \gamma_j \quad \forall i,j \in N, \label{eq:queue_st1}\\
    & \quad  \sum_{j} \alpha_{ij} = 1 \quad \forall i \in N,  \label{eq:queue_st2}\\    
    & \quad \alpha_{ij}, \psi_{i}\geq 0 \quad \forall i,j \in N,
\end{align}
\end{subequations}
\rev{where $\psi_i$ is the rebalancing rate from station $i$, $\alpha_{ij}$ the probability of rebalancing from $i$ to $j$, and the service rate is $\gamma_{i} = \pi_{i}/(\lambda_i+\psi_i)$, with $\lambda_i$ the passenger arrival rate and $\pi_i$ the throughput of the closed Jackson network. 
These constraints directly encode queueing-theoretic balance conditions.}

\subsubsection{Integration of matching and rebalancing}
\rev{Many works jointly optimize matching and rebalancing~\cite{guo2021robust,iacobucci2019optimization,rossi2018routing,tsao2018stochastic,carron2019scalable}, since both shape overall vehicle flows, as reflected in \cref{eq:reb_st2} where~$y_{ij}^k$ captures matched trips. With ridepooling, however, this joint problem becomes substantially more complex, adding another layer of difficulty to an already challenging assignment task (see \cref{sec:operation-all}). 
Consequently, a more common strategy is to treat matching and rebalancing in separate steps~\cite{alonso2017demand,alonso2017predictive,fielbaum2021demand}.}

\subsubsection{Matching optimization}
\rev{When matching is studied together with rebalancing, it is often assumed that passengers are matched with vehicles within the same region~\cite{gammelli2021graph,schmidt2024learning}, while rebalancing relocates vehicles between regions to serve future demand. 
When modeled separately, matching is typically formulated in a vehicle-centric fashion~\cite{liu2019dynamic,treleaven2013asymptotically,ruch2020the+}, leading to a bipartite assignment between vehicles and passengers.
It is often a \gls{abk:bpm} with canonical assignment formulation~\cite{motzkin1956assignment}, commonly solved using the Hungarian method~\cite{kuhn1955hungarian,bertsekas1993parallel}.}

\subsubsection{Future-state optimization via MPC}
\rev{Rebalancing and matching depend on both current and future system states. \gls{abk:mpc} addresses this by optimizing over a finite horizon using demand forecasts~\cite{guo2021robust,iacobucci2019optimization}, obtained from historical data~\cite{spieser2016shared,zgraggen2019model} or prediction models~\cite{he2023data,tsao2018stochastic,miao2017data,guo2022data}. Optimization-based methods enforce constraints and come with optimality guarantees, but \gls{abk:mpc} becomes less tractable as system size and problem complexity grow due to the combinatorial explosion of decisions.}

\subsubsection{Reproducibility considerations}
\rev{Much optimization-based \gls{abk:amod} research emphasizes scalability; claims in this direction should be supported by a clear explanation of how the formulation improves scalability, both theoretically and empirically~\cite{beiranvand2017best}. 
A key pitfall is \emph{implicit relaxation}, where the original problem is quietly simplified at implementation time to gain tractability. 
This obscures the true complexity and undermines fair comparison. 
All relaxations and approximations should therefore be stated explicitly and justified in, or immediately after, the problem formulation.}

\subsection{Learning} \label{sec:RL}
\rev{Learning-based methods, particularly \gls{abk:rl}, have recently gained traction for controlling \gls{abk:amod} systems, as they enable agents to learn decision strategies from experience in dynamic, uncertain environments~\cite{sutton1992reinforcement}. 
Formally, \gls{abk:rl} problems are typically modeled as \glspl{abk:mdp} with tuple~$(\mathcal{S}, \mathcal{A}, \mathcal{P}, d_0, R)$, where~$\mathcal{S}$ is the state space, $\mathcal{A}$ the action space,~$\mathcal{P}(s'|s,a)$ the transition kernel,~$d_0(s_0)$ the initial state distribution, and~$R: \mathcal{S} \times \mathcal{A} \to \mathbb{R}$ the reward function.
The goal is to learn a (typically stochastic) policy~$\pi(a|s)$ that maximizes expected discounted cumulative reward over horizon $T$. 
The probability of a trajectory~$\tau = (s_0,a_0,\ldots,s_T,a_T,s_{T+1})$ under $\pi$ is~$p_{\pi}(\tau) = d_{0}(s_0)\prod_{t=0}^{T}\pi(a_{t}|s_{t})\mathcal{P}(s_{t+1} | s_t,a_t)$, 
and the optimization problem is~$\textstyle{\underset{\pi}{\max} \quad \mathbb{E}_{\tau \sim p_\pi(\tau)} [\sum_{t=0}^{T}\gamma^{t}R(s_t,a_t)]}$, where $\gamma \in (0,1]$ discounts future rewards. 
In \gls{abk:amod} control, the state space typically includes demand (e.g., arrivals, queue lengths, travel times), fleet status, and network configuration, while actions comprise operational decisions such as rebalancing flows and vehicle–passenger matching. 
For a detailed survey of \gls{abk:rl}-based \gls{abk:amod} control, see~\cite{chouaki2025review}.}

\subsubsection{Online \glsentrytext{abk:rl}}
\rev{Most \gls{abk:rl} algorithms follow an online interaction loop: at each step the agent observes the state~$s_t$, selects an action~$a_t$, receives a reward~$r_t = R(s_t,a_t)$, and transitions to~$s_{t+1}$, using these trajectories to iteratively improve the policy. 
Two main paradigms are used in \gls{abk:amod}: (1) a centralized controller that observes the full system state and issues joint actions for all vehicles~\cite{jindal2018optimizing,gachter2021image}; and (2) decentralized control, where each vehicle acts as an independent agent with local information, i.e., \gls{abk:marl}~\cite{he2023robust,lin2018efficient}. Centralized control can better coordinate the fleet but suffers from high-dimensional state and action spaces at scale, whereas \gls{abk:marl} improves scalability by decomposing the problem but complicates training due to inter-agent interactions, making the design of communication and cooperation mechanisms a central research focus.}

\subsubsection{Offline \glsentrytext{abk:rl}}
Traditional (online) \gls{abk:rl} requires continual interaction with the environment or a simulator, \rev{which is often impractical in real-world \gls{abk:amod} systems due to safety, cost, or limited simulation fidelity.
Offline (or \emph{batch}) \gls{abk:rl} instead trains policies solely from pre-collected trajectories~\cite{levine2020offline}, avoiding active exploration but introducing challenges from distributional shift between the behavior policy that generated the data and the learned policy. 
This mismatch can cause compounding errors at evaluation time, and much of the literature focuses on mitigating these effects~\cite{li2025learning,schmidt2024learning}.}

\subsubsection{Imitation learning}
\rev{In offline \gls{abk:rl}, training uses pre-collected trajectories of state–action pairs. 
When these actions are known to be high quality, the agent can be trained in a supervised fashion to map states to actions, a process known as imitation learning~\cite{hussein2017imitation}. 
In \gls{abk:amod}, such guarantees typically come from optimization-based solutions or carefully designed heuristics~\cite{baty2024combinatorial}. 
While optimization models may offer optimality guarantees, they are often intractable for real-time control in large networks; imitation learning seeks to distill the structural patterns of these solutions into a policy that can be executed efficiently online, avoiding repeated optimization.}

\subsubsection{Reproducibility considerations}
\rev{For \gls{abk:rl}-based \gls{abk:amod} control, authors should explicitly specify the four core \gls{abk:mdp} components, state space, action space, reward function, and transition dynamics, detailing how demand, vehicle states, and control decisions are encoded. 
This clarifies how the original problem is cast as an \gls{abk:mdp} and enables fair comparison. 
The policy representation (often a neural network) must also be described, including architecture, hyperparameters, and optimization method. 
For offline \gls{abk:rl} and imitation learning, offline trajectories should be released or carefully documented, reflecting broader \gls{abk:rl} standards on data and implementation transparency~\cite{semmelrock2023reproducibility}. 
Finally, given the high dimensionality of \gls{abk:amod}, authors should report convergence evidence, robustness across random seeds, and performance across different system scales to support claims of real-world readiness.}

\subsection{Combination of optimization and learning}
\rev{Optimization and \gls{abk:rl} offer complementary strengths for \gls{abk:amod} control. 
Optimization-based methods provide interpretability, robustness, and performance guarantees but require repeatedly solving complex problems at each decision step, which can be computationally demanding at scale or in real time. \gls{abk:rl} shifts this burden to offline training, enabling fast online decisions with fewer modeling assumptions, but typically lacks strong guarantees and is harder to debug and interpret. 
Motivated by this complementarity, an increasing number of works integrate optimization and learning into hybrid approaches, broadly grouped into three categories.}

\subsubsection{Learning to guide optimization}
\rev{Here, the optimization structure is preserved, while learning is used to parametrize or simplify parts of the problem. 
For example, instead of directly learning \gls{abk:od}-level rebalancing flows, a \gls{abk:rl} agent can learn a target distribution of available vehicles across regions~\cite{gammelli2021graph,gammelli2022graph,tresca2025robo}; actual flows are then obtained by solving a lightweight optimization problem that minimizes rebalancing cost subject to this target. 
This reduces the action space from $\mathcal{O}(N^2)$ to $\mathcal{O}(N)$, improves scalability, enforces domain constraints, and facilitates offline \gls{abk:rl} by combining optimization modules with learned policies~\cite{schmidt2024offline}.}

\subsubsection{Learning to improve optimization objectives}
\rev{A second class of methods uses learning to enhance non-myopic optimization formulations by providing learned inputs such as weights, preferences, or forecasts. 
Examples include learning arc weights in bipartite matching graphs or cost-to-go estimates to guide multistep rebalancing decisions~\cite{liang2021integrated,enders2023hybrid,woywood2024multi,xu2018large,shah2020neural}. 
These approaches preserve the structure and guarantees of classical optimization while gaining flexibility and adaptability from the learned components.}
\begin{figure}[t]
\centering
\includegraphics[width=0.5\textwidth]{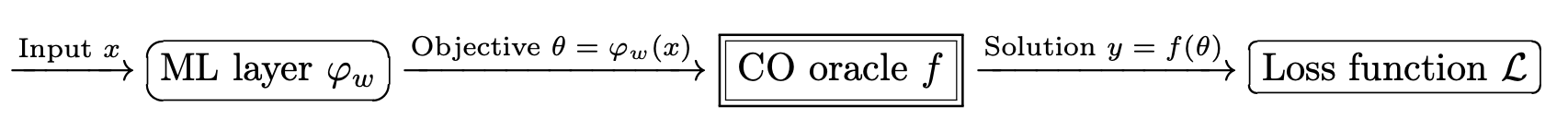}
\caption{CO-enriched ML pipeline \cite{dalle2022learning}.}\label{fig:co-ml}
\end{figure}

\subsubsection{Combinatorial optimization-enriched machine learning}
\rev{Beyond standard \gls{abk:rl}, recent work embeds \gls{abk:co} directly into \gls{abk:ml} pipelines, an approach termed \gls{abk:co}-enriched \gls{abk:ml}~\cite{dalle2022learning}. 
It is particularly useful when high-quality solutions are available offline from exact solvers or strong heuristics. 
The key idea is to train an \gls{abk:ml} model to predict parameters of a simplified \gls{abk:co} problem that is easier to solve yet preserves essential structure. 
As illustrated in \cref{fig:co-ml}, a scenario $x$ is mapped by an \gls{abk:ml} model to a parametrized \gls{abk:co} problem; a solver $f$ returns a solution $y$, and training minimizes the objective gap between $y$ and the ground-truth solution $\overline{y}$ via~$\mathcal{L}(\theta, \overline{y}) = \theta ^ Tg(y) +h(y) - (\theta ^ Tg(\overline{y}) +h(\overline{y}))$, where $\theta ^T g(y) + h(y)$ encodes the objective of the simplified problem.
A central challenge is the non-differentiability of combinatorial objectives, which are piecewise constant with vanishing gradients; probabilistic surrogate methods are therefore used to approximate gradients during training~\cite{dalle2022learning}. 
These techniques have shown promise in data-scarce regimes, including \gls{abk:amod} dispatching~\cite{baty2024combinatorial,jungel2023learning}. While both \gls{abk:rl} and \gls{abk:co}-enriched \gls{abk:ml} learn mappings from problem configurations (states) to operational decisions (actions), \gls{abk:rl} propagates signals through value functions or policies, whereas \gls{abk:co}-enriched \gls{abk:ml} learns to parametrize the optimization problem itself, often yielding more sample-efficient training when (near-)optimal solutions are known~\cite{dalle2022learning}. 
Recently, \gls{abk:co}-enriched \gls{abk:ml} has also been embedded within \gls{abk:rl} frameworks to combine the strengths of both paradigms~\cite{hoppe2025structured}.}

\subsubsection{Reproducibility considerations}
In hybrid approaches, reproducibility hinges on a clear specification of how optimization and learning are combined.
\rev{Key aspects to be documented include a) the structure and constraints of the optimization model, b) the role of learning in parameter estimation, objective shaping, or problem simplification, c) the model architecture, training dataset, and loss function (for learned components), d) the algorithm used for training, and e) evaluation metrics and baselines for comparison.
A well-documented allocation of computational resources between the modules is crucial to interpreting results and enabling reuse.}

\subsection{Heuristics}
\rev{Beyond the above paradigms, \gls{abk:amod} research also develops ad-hoc heuristics tailored to specific settings, e.g., localized dynamics, non-uniform demand, or tight computational budgets. 
For systems with advance reservations,~\cite{duan2020centralized} propose a two-stage scheme: short-term routing via optimization and long-term route design via a heuristic informed by these decisions. 
\cite{huang2024bi} introduce a hierarchical strategy with a high-level learning-based inter-regional rebalancer and a low-level coverage controller for intra-regional motion. 
To handle spatially correlated or bursty demand,~\cite{ruch2020the+} dispatch multiple vehicles to a single request. 
Other works use heuristics to accelerate optimization~\cite{levin2017general,simonetto2019real,li2021real,li2019efficient} or adopt rule-based strategies~\cite{fagnant2018dynamic,liu2019dynamic,maciejewski2016assignment,hyland2018dynamic} that reduce decision complexity in practice.}

\rev{Given their domain-specific nature, heuristic algorithms should clearly state their problem setting, assumptions, and intended use case to clarify contributions, contextualize performance, and indicate when they are applicable or extensible. 
A summary of key \gls{abk:amod} studies, including modeling frameworks, operational focus, constraints, and algorithmic approach, is provided in \cref{tab:summary}.}

\begin{table*}[t!]
    \centering
    \caption{Previous work on \gls{abk:amod} control with different methods.}
    \begin{threeparttable}
    \adjustbox{max width=\textwidth}{
    \begin{tabular}{|c|c|c||c|c|c||c|c|c||c|c|c|c||c|}
        \hline
        \multicolumn{3}{|c||}{\textbf{System model}} & 
        \multicolumn{3}{c||}{\textbf{Operations}} & 
        \multicolumn{3}{c||}{\textbf{Constraints}} & 
        \multicolumn{4}{c||}{\textbf{Algorithm}} &
        \multirow{2}{*}{\textbf{Reference}} \\
        \cline{1-13}
        \textbf{N} & \textbf{Q} & \textbf{V} & 
        \textbf{Matching} & \textbf{Rebalancing} & \textbf{Ridepooling} & 
        \textbf{Charging} & \textbf{Parking} & \textbf{Fleet} & 
        \textbf{O} & \textbf{L} & \textbf{Combined} & \textbf{Customized} & 
        \textbf{} \\
        \hline\hline
        \checkmark&  &   &\checkmark&  &  &  &  &  &\checkmark&  &  &  & \cite{pavone2012robotic} \\
        \hline
        \checkmark&  &  &  &\checkmark&  &  &  &  &\checkmark&  &  &  & \cite{rossi2018routing,volkov2012markov,salazar2019congestion,spieser2016shared,sadeghi2022reinforcement,iglesias2018data} \\
        \hline
        \checkmark&  &  &  &\checkmark&  &  &  &\checkmark&\checkmark&  &  &  & \cite{volkov2012markov} \\
        \hline
        \checkmark&  &  &\checkmark&\checkmark&  &  &  &  &\checkmark&  & & & \cite{solovey2019scalable, wollenstein2021routing, miao2017data,rossi2018routing,carron2019scalable,tsao2018stochastic} \\
        \hline
        &  &\checkmark&\checkmark&  &  &\checkmark&  &  &\checkmark&  & & & \cite{boewing2020vehicle} \\
        \hline
        &  &\checkmark&\checkmark&  &  & &  &  &  &  &\checkmark& & \cite{enders2023hybrid,hoppe2024global,xu2018large} \\
        \hline
        &&\checkmark&\checkmark&\checkmark&\checkmark& &  &  &\checkmark&&  & & \cite{alonso2017predictive,alonso2017demand,fielbaum2021demand} \\
        \hline
        \checkmark&&\checkmark&&\checkmark&& &  &  &&&&\checkmark& \cite{huang2024bi} \\
        \hline
        \checkmark& & & &\checkmark& & &  &  &\checkmark& & & & \cite{sayarshad2017non} \\
        \hline
        &\checkmark& & &\checkmark& &\checkmark&  &  &\checkmark& & & & \cite{iglesias2019bcmp} \\
        \hline
        \checkmark& & &\checkmark&\checkmark& & &  &  & & &\checkmark& & \cite{jungel2023learning} \\
        \hline
        & &\checkmark&\checkmark&\checkmark& & &  &  & & &\checkmark& & \cite{woywood2024multi} \\
        \hline
        \checkmark& & &\checkmark& &\checkmark& &  &  & & & &\checkmark& \cite{levin2017general,yu2019integrated} \\
        \hline
        & &\checkmark&\checkmark& &\checkmark& &  &  &\checkmark& & & & \cite{miller2017predictive} \\
        \hline
        &\checkmark& & &\checkmark& & &  &  &\checkmark& & & & \cite{zhang2018analysis,zhang2015queueing,zhang2016control} \\
        \hline
        &\checkmark& &\checkmark& & &\checkmark&  &  &\checkmark& & & & \cite{belakaria2019multi} \\
        \hline
        \checkmark& & &\checkmark&\checkmark&\checkmark& &  &\checkmark& & & &\checkmark& \cite{fagnant2018dynamic} \\
        \hline
        \checkmark& & &\checkmark& &\checkmark& &  &\checkmark&\checkmark& & & & \cite{li2019efficient} \\
        \hline
        \checkmark& & &\checkmark& & & &\checkmark& &\checkmark& & & & \cite{levin2017congestion} \\
        \hline
        & &\checkmark&\checkmark&\checkmark&\checkmark& &  & & & & &\checkmark& \cite{liu2019dynamic} \\
        \hline
        & &\checkmark& &\checkmark&\checkmark& &  & & &\checkmark& & & \cite{lin2018efficient} \\
        \hline
        \checkmark& & & &\checkmark&\checkmark& &  & &\checkmark& & & & \cite{chen2017hierarchical,tsao2019model} \\
        \hline
        & &\checkmark&\checkmark& & & &  & & & & &\checkmark& \cite{treleaven2013asymptotically,hyland2018dynamic, maciejewski2016assignment} \\
        \hline
        & &\checkmark&\checkmark&\checkmark& & &  & & & & &\checkmark& \cite{ruch2020the+,duan2020centralized} \\
        \hline
        \checkmark& & & &\checkmark& & &  &\checkmark&\checkmark& & & & \cite{spieser2016vehicle} \\
        \hline
        \checkmark& & &\checkmark& &\checkmark& &  & &\checkmark& & & & \cite{li2021optimal} \\
        \hline
        & &\checkmark&\checkmark&\checkmark& &\checkmark&  & &\checkmark& & & & \cite{iacobucci2019optimization,zhang2016model} \\
        \hline
        \checkmark& & & &\checkmark& & &  & & & & &\checkmark& \cite{albert2019imbalance} \\
        \hline
        & &\checkmark&\checkmark&\checkmark& & &  & & &\checkmark& & & \cite{han2016routing} \\
        \hline
        & &\checkmark& &\checkmark& & &  & & &\checkmark& & & \cite{gachter2021image,wen2017rebalancing} \\
        \hline
        & &\checkmark& &\checkmark&\checkmark&\checkmark&  & & & &\checkmark& & \cite{james2019online} \\
        \hline
        \checkmark& & &\checkmark& & &\checkmark&  & &\checkmark& & & & \cite{he2023data} \\
        \hline
        \checkmark& & & &\checkmark& &\checkmark&  & & &\checkmark& & & \cite{he2023robust} \\
        \hline
        & &\checkmark& &\checkmark&\checkmark& &  & &\checkmark& & & & \cite{wallar2018vehicle} \\
        \hline
        & &\checkmark&\checkmark& &\checkmark& &  & & & &\checkmark& & \cite{kim2024learning} \\
        \hline
        & &\checkmark&\checkmark& &\checkmark& &  & & &\checkmark& & & \cite{shah2020neural,jindal2018optimizing} \\
        \hline
        & &\checkmark&\checkmark&\checkmark&\checkmark& &  & & & & &\checkmark& \cite{simonetto2019real} \\
        \hline
        \checkmark& & &\checkmark&\checkmark& & &  &\checkmark&\checkmark& & & & \cite{kang2021maximum} \\
        \hline
        & &\checkmark&\checkmark&\checkmark&\checkmark&\checkmark&\checkmark&  &\checkmark& & & & \cite{chu2021joint} \\
        \hline
        & &\checkmark&\checkmark&\checkmark&\checkmark& &\checkmark&\checkmark  &\checkmark &\checkmark& & & \cite{ma2017designing} \\
        \hline
        & &\checkmark&\checkmark& & &\checkmark& & & & & &\checkmark& \cite{li2021real} \\
        \hline
        \checkmark& & & &\checkmark& &\checkmark& &\checkmark&\checkmark& & & & \cite{guo2020vehicle} \\
        \hline
        & &\checkmark&\checkmark&\checkmark&\checkmark& & & & &\checkmark& & & \cite{gueriau2020shared} \\
        \hline
        & &\checkmark&\checkmark&\checkmark& & & & &\checkmark& & & & \cite{dandl2019evaluating} \\
        \hline
        \checkmark& & & &\checkmark& & & & & & &\checkmark& & \cite{gammelli2021graph} \\
        \hline
        & &\checkmark& &\checkmark& &\checkmark& & & & & &\checkmark& \cite{hu2020artificial} \\
        \hline
    \end{tabular}
    }
    \begin{tablenotes}
        \item[1] In the system model column, ``N'' represents ``Network flow'', ``Q'' represents ``Queue-theoretical'', and ``V'' represents ``Vehicle-centric''.
        \item[2] In the constraints column, ``Fleet'' means that the number of vehicles appears as a constraint or decision variable in the model explicitly.
        \item[3] In the algorithm column, "O" represents ``Optimization'' and ``L'' represents ``Learning''.
    \end{tablenotes}
    \end{threeparttable}
    \label{tab:summary}
\end{table*}

%% file: chapters/assumption.tex
\section{Modeling assumptions} \label{sec:assumption}
\rev{Assumptions play a central role in shaping \gls{abk:amod} algorithms by completing the model and delimiting their intended use cases. This section reviews common assumptions in \gls{abk:amod} research to highlight gaps between experimental setups and real-world deployments.
Clearly specifying these assumptions is essential for meaningful benchmarking.}

\subsection{Modeling demand}
Passenger demand plays a crucial role in shaping operational decisions, as discussed in \cref{sec:algo}.
Various assumptions can be made about demand behavior, each influencing the modeling approach and decision-making process.
\subsubsection{Time-varying vs. time-invariant} 
\rev{\emph{Time-varying} demand fluctuates over time due to travel patterns or elastic effects induced by \gls{abk:amod} operations (e.g., dependence on vehicle availability), whereas \emph{time-invariant} demand is assumed constant. 
The latter is common in queueing-theoretical models to estimate expected waiting times and derive equilibrium conditions~\cite{iglesias2019bcmp,zhang2016control}, and is reasonable when demand evolves slowly, as often observed in dense urban settings~\cite{neuburger1971economics}.}
\subsubsection{Deterministic vs. stochastic}
\rev{Deterministic demand is assumed precisely known, either directly from historical data or via accurate forecasts. 
Stochastic demand instead models passenger arrivals as a random process, typically Poisson~\cite{pavone2012robotic,hyland2018dynamic,guo2020vehicle,volkov2012markov}, an assumption ubiquitous in queueing-theoretical work where arrivals form part of a queueing system. 
These demand assumptions parallel those in power grid and communication network control, and making them explicit is equally important for reproducibility.}

\subsection{Accounting for congestion}
\rev{Congestion is a key aspect of transportation management, driving both emissions and lost productivity~\cite{schrank20112011}. 
Most \gls{abk:amod} control studies treat congestion as \emph{exogenous}, assuming fixed travel times independent of fleet operations, which is reasonable when \gls{abk:amod} penetration is low~\cite{Pavone2015}. 
As fleets grow, however, \gls{abk:amod} operations can substantially contribute to congestion~\cite{oh2020assessing,oh2021impacts}, requiring \emph{endogenous} models where fleet activity affects network dynamics.}
\subsubsection{Threshold models}
\rev{Vehicles travel at free-flow speed as long as flows remain below a critical threshold on each segment; beyond this, flow constraints are imposed~\cite{iglesias2019bcmp,salazar2018interaction}. Rooted in classical traffic theory~\cite{wardrop1952road}, this is suitable when the goal is to prevent, rather than fully model, congestion.}
\subsubsection{BPR Model}
The \gls{abk:bpr} model is the most widely used method to model congestion in \gls{abk:amod} systems~\cite{salazar2019congestion,solovey2019scalable,wollenstein2021routing,zgraggen2019model}. 
Developed in~\cite{us1964traffic}, it \rev{relates traffic flow and travel time}:
\begin{equation}
    t_{a}(x_{a}) = t_{a}^{0} \bigg(1 + 0.15 \Big(\frac{x_a}{\gamma_a} \Big)^4 \bigg),\label{eq:bpr}
    \end{equation}
\rev{where $t_{a}^{0}$ is free-flow travel time, $x_a$ the flow, and $\gamma_a$ the capacity of road $a$. 
Tracking link flows thus yields congestion-dependent travel times with low computational cost.}

\subsubsection{MFD}
\rev{The \gls{abk:mfd} characterizes the relationship between flow, density, and speed at the network level~\cite{huang2024bi}. 
Given a link flow, the \gls{abk:mfd} provides a corresponding speed (and thus travel time). 
Unlike \gls{abk:bpr}, an \gls{abk:mfd} must be calibrated for each specific network using link and intersection parameters~\cite{daganzo2008analytical}.}

\subsubsection{Cell and Link Transmission Models}
\rev{The \gls{abk:ctm} models traffic along a road by discretizing it into cells~\cite{daganzo1994cell,levin2017general,lazar2021learning,levin2017congestion}; congestion propagates from bottlenecks through the cells, dynamically reducing speeds. 
The \gls{abk:ltm} is a coarser variant that tracks flows only at link boundaries, reducing computational effort~\cite{yperman2005link}.}

\rev{Most current \gls{abk:amod} control policies ignore endogenous congestion, arguing low service penetration; yet as \gls{abk:amod} shares grow, these effects become non-negligible. \gls{abk:amod} can increase total vehicle-kilometers and congestion via rebalancing and deadheading~\cite{oke2020evaluating}, with associated carbon emissions. Transport is a rapidly growing source of global emissions, and in the U.S. about 80\% of transport-sector emissions stem from individual ground travel~\cite{transportation_decarbonization}. 
Future work should explicitly model the impact of \gls{abk:amod} on road usage and design control algorithms that constrain congestion and emissions.} 

\subsection{Modeling passengers}
It is important to model how passengers respond to the performance of the system, specifically in relation to demand elasticity.
When demand is \emph{inelastic}, passengers remain in the system until they are matched with a vehicle.
However, in multimodal transportation systems, where alternative travel options exist, passengers can abandon the system if they experience excessive waiting time or increased trip costs.
\rev{\emph{Elastic} demand can be modeled in various ways.}

\subsubsection{Passenger loss model}
Here, passengers are sensitive to time and exit the system if they are not matched within the same time window in which they requested a trip~\cite{boewing2020vehicle,miller2017predictive,xu2018large,pavone2012robotic}.
This is called a rejection model, where passengers leave if their request is denied~\cite{enders2023hybrid, woywood2024multi}, and this class of models is particularly relevant in systems requiring a high level of service reliability.

\subsubsection{Maximum passenger waiting time}
Here, passengers remain in the system until their waiting time exceeds a predefined maximum waiting time or designated pickup interval~\cite{he2023robust,hoppe2024global,gueriau2020shared,hu2020artificial,wen2017rebalancing,spieser2016shared}.
Otherwise, they abandon the system.

\subsubsection{Maximum delay model}
Passengers are only matched if their trip can be completed before a specified drop-off time~\cite{alonso2017demand,shah2020neural,simonetto2019real}.
Otherwise, they remain in the system until their latest acceptable pickup time expires.

\subsubsection{Choice model}
Passengers make travel decisions based on utility functions that incorporate factors such as waiting time, cost, and service quality~\cite{kim2024learning,liu2019framework,kim2024estimate}.
This model provides a more realistic approximation of human decision-making and allows for a probabilistic representation of passenger behavior.
Integrating choice models into \gls{abk:amod} control modules is an emerging and promising research direction. When applied, the elasticity of the choice to the traveling attributes, such as the value of time, needs to be reported to justify the rationality of the passenger behavior, and to clarify the scope of the impact of the control actions.

\subsection{Unknown demand}
\rev{Because future demand is unknown yet crucial for non-myopic decisions, \gls{abk:amod} control relies on several strategies. 
A common approach is to exploit historical data, either by using past demand as a proxy for future demand or by training predictive models. 
This requires high-quality data and accurate forecasts, and even then predictions may be biased, since historical demand reflects historical supply. In such cases, censored demand forecasting is needed to obtain unbiased estimates of latent demand~\cite{gammelli2020estimating}.}

\rev{To reduce reliance on accurate forecasts, robust optimization can be used to hedge against uncertainty, optimizing decisions for worst-case demand distributions~\cite{he2023robust,guo2021robust} or uncertainty sets~\cite{miao2017data}. 
When future demand (or its distribution) is hard to model, two further options are common: repeatedly solving a deterministic optimization problem online as new requests arrive (real-time optimization), or using \gls{abk:rl}, which learns policies that maximize expected return under stochastic demand without requiring explicit knowledge of underlying dynamics.}

\subsection{Reproducibility considerations}
The assumptions mentioned above can often be inferred from the model and experiment sections, even if they are not explicitly stated. For instance, the congestion assumption and the passenger model are sometimes important components that are integrated as constraints within the system model and algorithm design, and the demand assumption can be inferred for the experiment setup. However, we encourage that these assumptions be explicitly clarified in the problem definition. 
\rev{Doing so allows the reader to quickly identify the application scenario of the research work without needing to thoroughly review the entire work, thereby facilitating benchmarking.}

%% file: chapters/evaluation.tex
\section{Evaluation criteria}
\label{sec:evaluation}
\rev{The previous sections covered methodological aspects of \gls{abk:amod} research. 
Before running experiments, however, it is essential to define clear \emph{evaluation criteria}, as these determine how methods are benchmarked and contributions are demonstrated. 
This section focuses on benchmarking criteria specific to \gls{abk:amod} control; general benchmarking guidelines for other systems are discussed in~\cite{kounev2020systems}.}
\subsection{Computational complexity}
\rev{Controlling \gls{abk:amod} systems is computationally taxing due to their scale and rich interaction structure, making complexity a key determinant of real-world applicability and improvement over the state of the art.
Common comparison metrics include absolute solution times~\cite{solovey2019scalable,duan2020centralized,james2019online}, real-time optimality (solution quality under operational time limits)~\cite{alonso2017predictive,alonso2017demand,wollenstein2021routing}, convergence behavior for learning-based methods~\cite{hoppe2024global,woywood2024multi}, and scalability analysis with respect to problem size and computational complexity~\cite{levin2017congestion,treleaven2013asymptotically}.}
\subsection{Performance on control objectives}
\rev{A natural way to evaluate an algorithm is by measuring performance on its stated control objectives (\cref{sec:algo}), using metrics such as average travel time, distance, service rate, and revenue. 
These metrics should be consistent with model assumptions. 
For instance, if a maximum waiting time or service window is imposed, average waiting time alone may not reflect service quality; service rate under the imposed cap should also be reported. 
In models without strict waiting limits, long-wait outliers may be obscured by averages, raising equity concerns; reporting tail statistics (e.g., 95th-percentile waiting time~\cite{bischoff2016simulation}) helps reveal such disparities.}

\rev{Beyond efficiency gains, it is important to demonstrate \emph{robustness} across scenarios to support claims of generalization. This can be assessed via sensitivity analysis, varying demand patterns~\cite{wen2017rebalancing,gammelli2021graph}, algorithm-specific parameters (e.g., \gls{abk:av} penetration~\cite{yang2020planning}, capacity~\cite{simonetto2019real}), and fleet size~\cite{duan2020centralized,hyland2018dynamic}. 
Reporting how performance changes under such variations is crucial both for interpreting results and for reproducibility.}

\subsection{Externalities}
\rev{As \gls{abk:amod} adoption grows, its impact on the broader transport network becomes significant. 
A common criticism is that it can increase road congestion; an important contribution of new methods is therefore to show how they limit traffic impact while maintaining service quality, e.g., by reporting road utilization~\cite{iglesias2019bcmp,levin2017general,zhang2016control,salazar2018interaction} and travel times~\cite{lazar2021learning,rossi2018routing,gueriau2020shared}.}
\rev{A second, underexplored externality is equity. 
In \gls{abk:amod} control, equity does not mean identical service for everyone, but allocating resources according to the needs and constraints of different social groups~\cite{litman2017evaluating}. Equity impacts depend on factors such as service availability and affordability. 
As discussed in \cref{sec:evaluation}, most control objectives emphasize time- and profit-based metrics; without explicit equity goals, algorithms will naturally favor short, high-margin trips. 
Existing MoD work has largely focused on \emph{measuring} equity via tailored metrics~\cite{kagho2024framework,bhuyan2019gis}. 
A key next step for the \gls{abk:amod} community is to move toward \emph{equity-centric} control algorithms that optimize service quality and fairness jointly.}

%% file: chapters/experiment.tex
\section{Experiments} \label{sec:experiment}
The design of experiments plays a central role in validating proposed methods and is a cornerstone of reproducible research.
Reproducibility hinges on the replicability of experimental results and the validity of the conclusions drawn from them, which in turn requires a well-structured and transparently documented experimental setup.
In this section, we identify and discuss key components of experimental design that influence reproducibility in \gls{abk:amod} control research.
In addition, we provide a curated list of publicly available datasets and simulators to support standardized experimentation and facilitate future benchmarking efforts.
\subsection{Demand datasets}
\rev{Trip record data is the most commonly used data type in \gls{abk:amod} control research, typically including pickup and dropoff times and locations, travel distances, and in some cases, additional metadata such as fare amounts and payment types.}
Several publicly available datasets have become de facto standards in the field, supporting a range of simulation and validation efforts.

\subsubsection{NYC Taxi \& Limousine Commission (TLC) Data}
This is the most widely used dataset in \gls{abk:amod} research~\cite{nyc_tlc_trip_data}.
\rev{It provides detailed trip-level data for the five boroughs of NYC, with Manhattan being the most commonly studied.}
The location data is anonymized using predefined taxi zones with the corresponding shapefiles accessible through the TLC website.
The TLC dataset is particularly valued for its rich metadata and well-maintained documentation, which reduces the burden on researchers to describe preprocessing steps in detail.
Moreover, the dataset is updated monthly, ensuring continued relevance and minimizing concerns about data obsolescence. 

\subsubsection{NYC Taxi Trip Data} 
Another variant of the NYC data is hosted by the Illinois Data Bank~\cite{illinoisdatabankIDB-9610843}.
This collection contains taxi trip records from 2019 to 2013, which were originally acquired from the NYC TLC.
\rev{It includes raw GPS coordinates for pickup and dropoff locations, was used in~\cite{donovan2015using}, and the accompanying preprocessing scripts are publicly available~\cite{donovan_published_code}. It offers a reproducible analysis pipeline for researchers working with GPS-based mobility data.}

\subsubsection{San Francisco Taxi Trace Data}
This dataset~\cite{c7j010-22} provides GPS traces for approximately 500 taxis from May to June 2008 in San Francisco.
It includes the taxi occupancy labels, which enable reconstruction of trips from continuous trace data.
This dataset has been leveraged in several \gls{abk:amod} studies~\cite{ruch2020the+,albert2019imbalance,tsao2019model}, especially to evaluate algorithms in scenarios with spatial heterogeneity and demand imbalance.

\subsubsection{Sioux Falls Data}
A more synthetic dataset is based on the Sioux Falls transportation network, included in a classical transportation benchmark database~\cite{transportation_networks}.
This dataset provides static \gls{abk:od} flow data without temporal resolution, originally intended for studying traffic assignment.
Despite its simplicity and lack of realism, the Sioux Falls network has been used in several \gls{abk:amod} control papers~\cite{liu2019dynamic,kang2021maximum}, often for debugging or testing algorithms in a controlled environment due to its small size and clean topology.

\subsubsection{Reproducibility considerations}
\begin{figure}[t]
\centering
\includegraphics[width=0.4\textwidth]{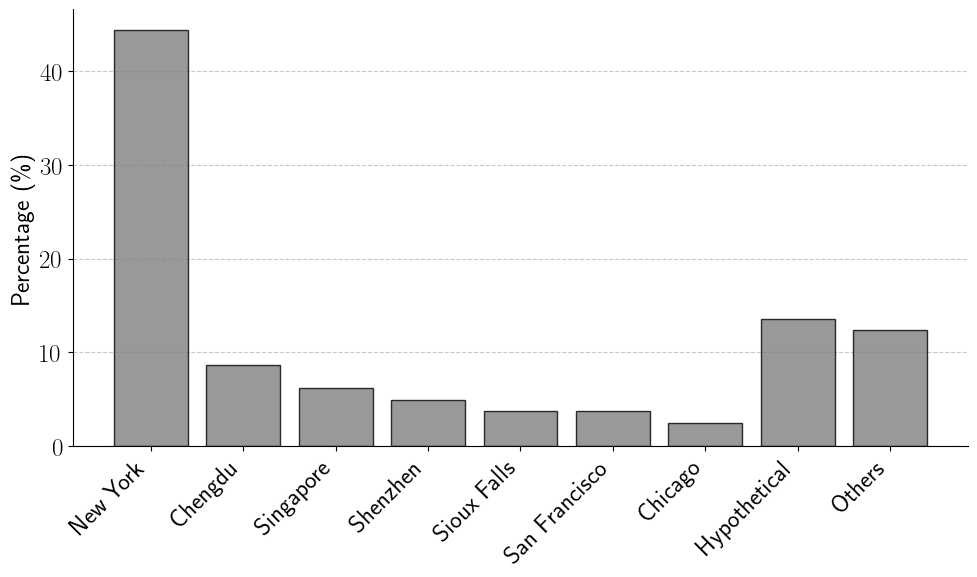}
\caption{Distribution of networks used in \gls{abk:amod} case studies.}\label{fig:scenario}
\end{figure}

\rev{To support reproducibility and cross-paper comparability, researchers are strongly encouraged to use publicly available, actively maintained datasets. 
\cref{fig:scenario} summarizes the networks used in the \gls{abk:amod} case studies we reviewed; New York is by far the most common, largely because the NYC TLC data are public and continuously maintained. 
This dataset plays a role for \gls{abk:amod} similar to ImageNet~\cite{imagenet_cvpr09} and KITTI~\cite{geiger2012we} in vision and robotics. 
We also provide tool to generate city-scale \gls{abk:amod} scenarios from these datasets at~\url{https://anonymous.4open.science/r/scenario-generation-E25A}.}

\rev{By contrast, although the Shenzhen network is used in many works, its data originate from a DiDi challenge~\cite{didiglobal2025} released in 2016 and are no longer publicly maintained or available at the source. 
Many other papers rely on proprietary city networks or closed datasets (grouped as ``others'' in~\cref{fig:region})~\cite{sadeghi2022reinforcement,braverman2019empty,lin2018efficient,he2023data,huang2024bi,li2021real}. 
When data cannot be shared, authors should at least provide detailed metadata, spatial coverage, horizons, demand distribution, sampling frequency, so others can construct comparable scenarios.}

\rev{Data preprocessing is another critical but often underreported step. Cleaning, sampling, trip filtering, demand aggregation, and especially network modifications (e.g., graph pruning to speed up computation) can strongly affect results. 
Any deviation from the raw dataset should therefore be documented, including thresholding, zone aggregation, link removal, trajectory simplification, and noise filtering. 
These details are frequently omitted yet are essential for fair comparison and replication.}

\subsection{Network granularity}
In real-world transportation systems, traffic flows along road links that connect intersections within a detailed urban network.
The number of such links and the complexity of their interactions significantly influence the scale of the \gls{abk:amod} control problem.
To manage computational complexity in experimental setups, researchers often aggregate road-level details into higher-level abstractions, typically in the form of regions or stations, where vehicles are assumed to move between such regions rather than individual road segments.
The \emph{granularity} of the network representation used in experiments is a critical design choice.
It affects both the fidelity of the simulation and the scalability of the proposed algorithm~\cite{tresca2025robo}.

\begin{figure}[tb]
\begin{center}
\begin{subfigure}[tb]{0.33\linewidth}
\includegraphics[width=\linewidth]{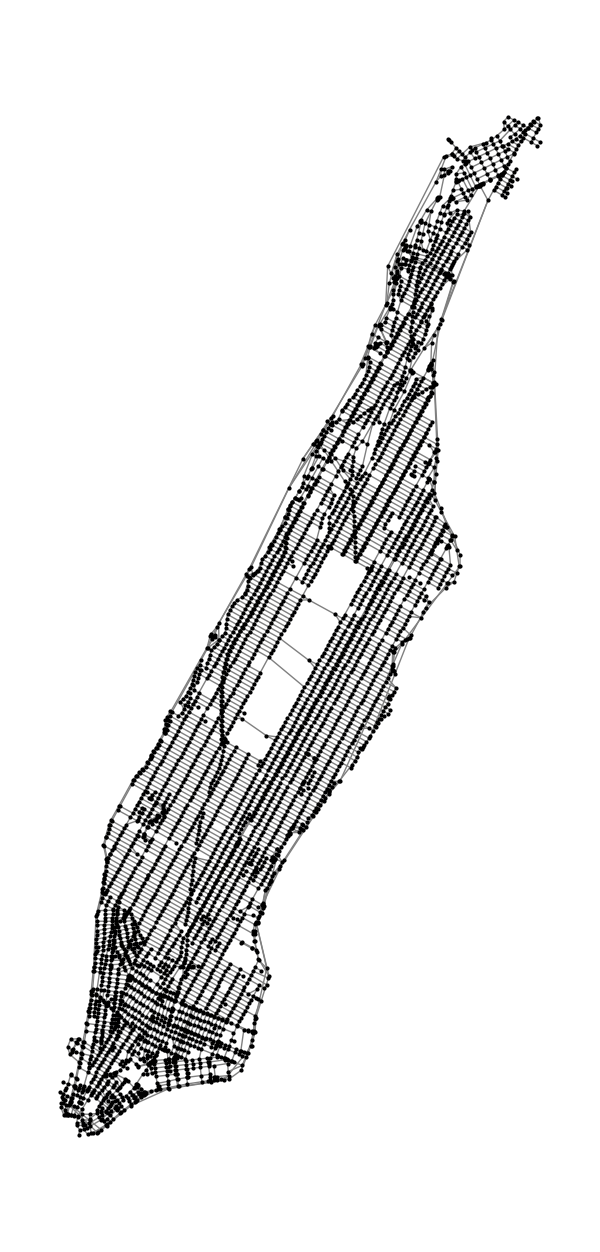}
\caption{Node-based. \label{fig:node}}
\end{subfigure}
\begin{subfigure}[tb]{0.33\linewidth}
\includegraphics[width=\linewidth]{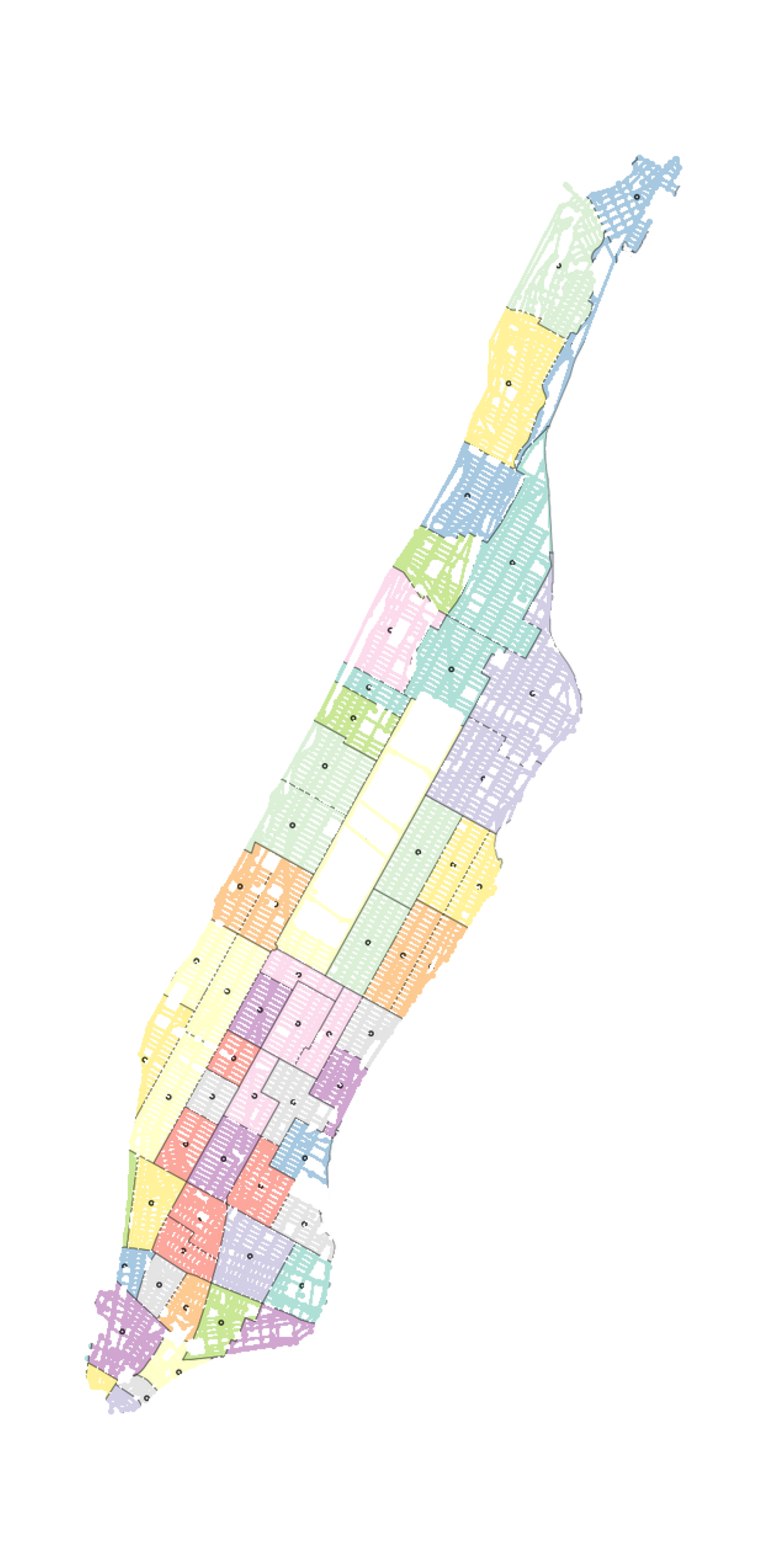}
\caption{Mixed. \label{fig:mix}}
\end{subfigure}
\begin{subfigure}[tb]{0.3\linewidth}
\includegraphics[width=\linewidth]{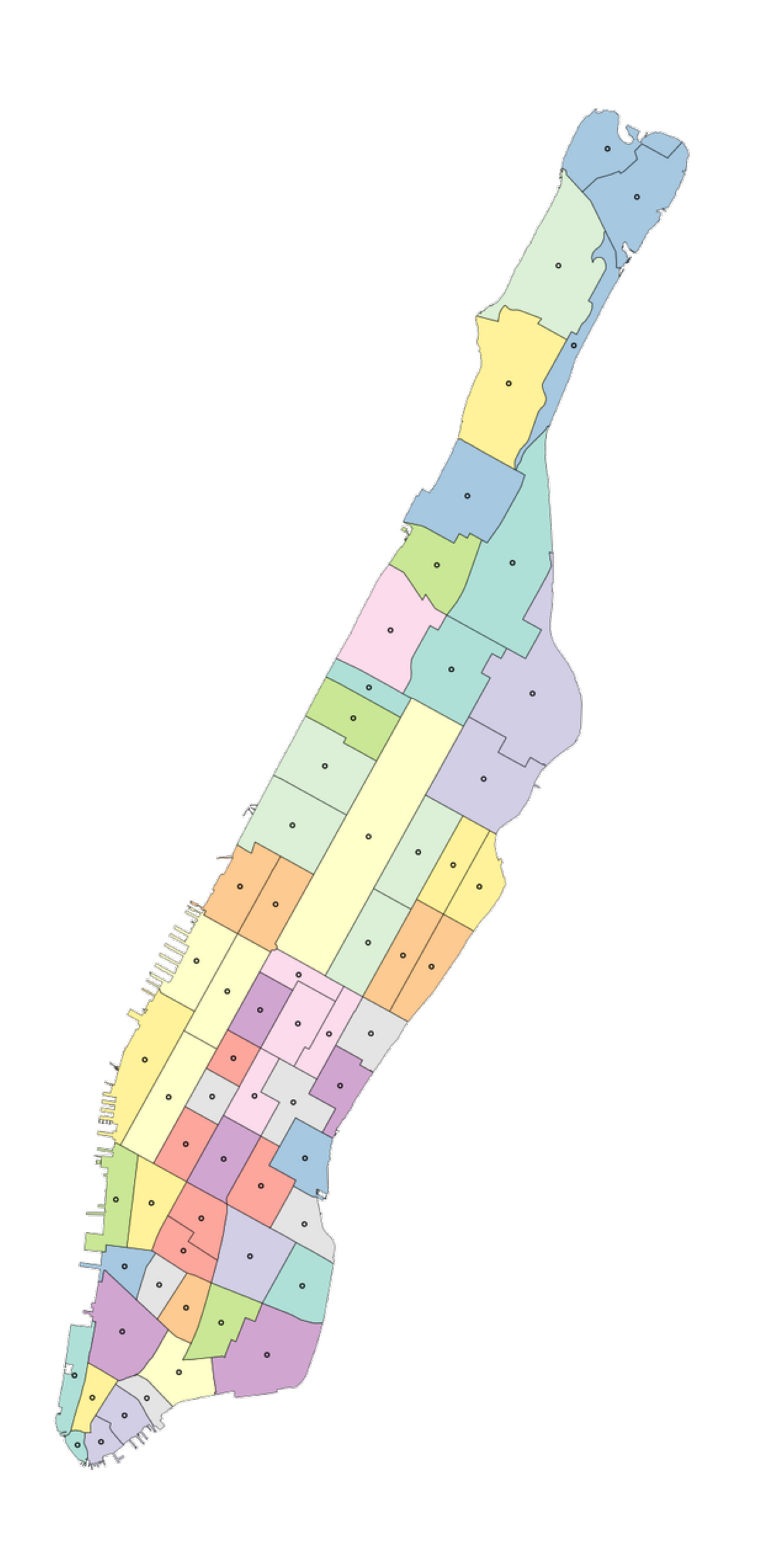}
\vspace{1pt}
\caption{Region-based. \label{fig:region}}
\end{subfigure}
\end{center}
\caption{Different network granularity for NYC.}
\label{fig:granularity}
\end{figure}

As the example illustrated in \cref{fig:granularity}, existing studies vary considerably in how they model the underlying network.
Such variations can be broadly categorized into three levels of decreasing granularity.
The highest level of detail is the \emph{node-based network} (\cref{fig:node}), where operations are modeled on an explicit road network with individual nodes and links~\cite{alonso2017demand,liang2021integrated,fielbaum2021demand}.
This approach offers the most realistic representation of urban topology and is particularly suitable for high-fidelity simulations.
However, it also introduces significant computational demands, making algorithmic efficiency a critical concern.
A more abstract formulation is the \emph{mixed network model} (\cref{fig:mix}), where decision-making, such as dispatching or rebalancing, is conducted at the region level, while vehicle movements are still simulated on the full node-link network~\cite{iglesias2018data,huang2024bi}.
This hybrid approach is common in rebalancing research, where idle vehicles can be repositioned without a fixed destination.
It balances operational realism with tractable decision-making, making it a practical compromise between accuracy and scalability.
Finally, the most simplified representation is the \emph{region-based network} (\cref{fig:region}), in which both operations and vehicle movements are modeled at the region level.
Pickup and dropoff locations are aggregated, and intra-region travel dynamics are ignored.
This abstraction is the most commonly used in \gls{abk:amod} control research due to its simplicity and reduced computational overhead, making it ideal for evaluating high-level policies or conducting large-scale sensitivity analysis.

In region-based experiments, the method used to partition the network into regions plays a critical role in determining the spatial resolution of the simulation.
Despite its importance, this aspect is often underreported or completely omitted in many published papers.
The number and layout of regions directly influence the fidelity of the experiment, the granularity of operational decisions, and ultimately, the perceived performance of the control algorithm.
Typically, three main approaches are used to define regional partitions.
The first approach leverages administrative or census regions defined by government agencies~\cite{sayarshad2017non}.
These partitions offer a consistent and policy-aligned geographical structure, but may not align with actual demand patterns or travel behavior.
The second approach clusters \gls{abk:od} pairs of trip demand using unsupervised learning techniques, most commonly k-means clustering~\cite{rossi2018routing,spieser2016vehicle,iacobucci2019optimization,albert2019imbalance}.
In these cases, it is important to explicitly specify how the number of clusters (i.e., regions) is chosen.
This number may be derived from external sources, such as the number of postal codes in a city~\cite{volkov2012markov}, or optimized internally, for example, to ensure that all nodes within a region are reachable from the center of the region~\cite{wallar2018vehicle}.
The third approach discretizes the network into a uniform grid of fixed-size cells~\cite{lin2018efficient,han2016routing,jindal2018optimizing}.
This method is straightforward and facilitates implementation, particularly in simulations that require efficient spatial indexing or integration with raster-based data.
However, the size and shape of the grid significantly affect the resolution and realism of the experiment and should always be reported explicitly.

Choosing the appropriate level of granularity involves trade-offs between modeling realism and computational efficiency.
\rev{As such, it is essential for researchers to clearly report the granularity used in their experiments and justify it in relation to the intended scope, evaluation metrics, and algorithms.}

\subsection{Simulators}
Since \gls{abk:amod} control strategies are designed to optimize system performance over extended time horizons, it is essential to simulate how control actions influence system dynamics over time.
The simulator serves as the backbone of any experimental pipeline, yet in much of existing \gls{abk:amod} research, simulation details are poorly documented or not made publicly available~\cite{riehl2025revisiting}.
This lack of transparency poses a major barrier to reproducibility.
Clear documentation of the simulation environment, including assumptions, parameters, and state transition logic, is critical to allow replicability and facilitate meaningful comparisons between studies.

Broadly, simulation in \gls{abk:amod} research takes two forms: custom-built simulators tailored to specific experimental setups, and general-purpose transportation simulators that offer broader functionality and higher fidelity.
When developing a custom simulator, researchers should provide a detailed description of the simulation pipeline, including how vehicle states, passenger behavior, and network evolution are modeled over time.
Good examples of this practice include~\cite{guo2021robust,levin2017general}, where authors use flowcharts and dedicated sections to clearly communicate their simulation framework. Another good practice of a well-documented simulator is in ~\cite{engelhardt2022fleetpy}, where the authors provide a detailed documentation of the different modules and available functionality of the developed \gls{abk:amod} simulator. The source code and user guide is also provided.
In contrast to the custom-built simulators, general-purpose simulators, developed and maintained by large research communities, offer mature infrastructure and extensive documentation.
These tools incorporate a wide range of features and assumptions, and they are often validated across multiple use cases.
The most commonly used general-purpose simulators for \gls{abk:amod} research are detailed in the following.

\subsubsection{MATSim~\texorpdfstring{\cite{w2016multi}}{}}
MATSim is an open-source, activity-based microscopic simulator designed for large-scale, agent-based transportation modeling.
It simulates individual traveler behavior using a queue-based traffic flow model and assigns plans to agents based on a scoring function interpreted as a utility model.
While MATSim does not explicitly simulate detailed car-following behavior, it efficiently captures congestion effects at scale.
Importantly, MATSim allows for integration of custom \gls{abk:amod} control algorithms by overriding the plan-choice module.

\subsubsection{SimMobility~\texorpdfstring{\cite{adnan2016simmobility}}{}}
SimMobility offers a multi-scale simulation framework capable of modeling interactions across short-, mid-, and long-term horizons.
Its short-term module captures microscopic behaviors such as lane-changing and braking;
the mid-term module focuses on route and mode choice dynamics;
and the long-term module simulates structural trends such as land use, vehicle ownership, and job location decisions.
Although SimMobility has been used in \gls{abk:amod} control~\cite{marczuk2016simulation}, it is more commonly employed to assess the systemic impact of \gls{abk:amod} systems on accessibility~\cite{nahmias2021evaluating,zhou2021simulating} and interactions with other transportation modes~\cite{oh2020assessing,nguyen2023examining}.
Its ability to capture behavioral adaptations and cross-modal feedback makes it particularly well-suited for urban planning and policy evaluation.

\subsubsection{SUMO~\texorpdfstring{\cite{lopez2018microscopic}}{}}
SUMO is another open-source microscopic simulator that emphasizes usability and extensibility.
Similar to MATSim, it supports large-scale simulation with relatively lightweight computational demand.
A key advantage of SUMO is its rich suite of auxiliary tools for building simulations based on real-world data.
It also features a \gls{abk:gui} for visualizing simulations and debugging control logic in real time.
The availability of Python bindings further improves SUMO's versatility, allowing it to be embedded into broader research pipelines and enabling easy integration with optimization or learning algorithms~\cite{tresca2025robo}.

\subsubsection{AMoDeus~\texorpdfstring{\cite{ruch2018amodeus}}{}}
This is a simulator specifically designed for \gls{abk:amod} systems.
Built on top of MATSim, it provides a unified framework for simulating and benchmarking a variety of fleet control algorithms.
AMoDeus includes a curated library of state-of-the-art AMoD policies, including support for ridepooling and electric fleet variants.
A comprehensive list of available algorithms is maintained in~\cite{amodeus_repository}, making AMoDeus one of the most reproducibility-oriented platforms currently available for AMoD experimentation.

When selecting or developing a simulator, it is important to consider the trade-off between fidelity and transparency.
General-purpose simulators offer realism and robust documentation, but may be harder to tailor for specific \gls{abk:amod} experiments.
Custom simulators allow for fine-grained control and rapid iteration but require meticulous documentation to support reproducibility.
Regardless of the choice, future research should prioritize clear communication of simulation assumptions, modularity in design, and public access to code and scenario definitions.

\subsection{Parameter values}
Running controlled experiments requires careful specification of both simulator and algorithm parameters.
Among these, one of the most impactful, yet frequently underreported, parameters is the fleet size.
The number of vehicles in the system directly affects system performance metrics such as waiting time, vehicle utilization, and rebalancing efficiency.
Consequently, the choice of fleet size should not be arbitrary; it should be motivated by clearly defined criteria and explicitly stated in the experimental setup.

Different studies adopt varying strategies for determining fleet size.
Some aim to provide enough vehicles to meet all demand~\cite{levin2017general}, others target performance guarantees such as limiting maximum passenger waiting time~\cite{fagnant2018dynamic}, or ensure a buffer of idle vehicles for effective rebalancing~\cite{gachter2021image}.
A more robust and widely adopted approach is to evaluate algorithm performance across a range of fleet sizes~\cite{iglesias2019bcmp,jungel2023learning,woywood2024multi,iacobucci2019optimization}.
This not only demonstrates the algorithm's scalability and generalizability, but also helps identify regimes in which it is most effective.
The chosen fleet size, when considered jointly with the scale and structure of the experimental dataset, provides important insights into the applicability of the proposed algorithm.
Reporting both allows readers to understand the operational context and assess the potential for deployment in real-world scenarios.
Moreover, it signals the algorithm's scalability and suitability for extension in future research.

In addition to fleet size, many other parameters, such as cost coefficients, vehicle speed, demand elasticity, and algorithm-specific hyperparameters, play a critical role in shaping the behavior and outcomes of the system.
These values should be presented systematically, ideally in a parameter table that includes i) the parameter's name, ii) the value used in the experiment, and iii) the rationale for its selection.
Providing this level of detail serves two key purposes: it supports reproducibility by allowing other researchers to replicate results exactly, and it enables adaptability by helping others transfer the approach to new datasets or scenarios with confidence.
Thorough documentation of parameter settings is thus not just good practice. It is essential for building a credible and reusable body of work in \gls{abk:amod} research.

\subsection{Randomness and Robustness}
When randomness is present in an algorithm or simulator, results can vary across different runs, even under identical experimental setups.
This is particularly true for data-driven control algorithms that incorporate \gls{abk:ml}, where sensitivity to training conditions and stochastic components (e.g., initialization, sampling, exploration) can lead to significant performance fluctuations~\cite{semmelrock2023reproducibility}.
To ensure the credibility and reproducibility of results, it is essential to report performance statistics over multiple independent experiments.
In particular, metrics such as the mean and standard deviation across random seeds provide a quantitative measure of the algorithm's robustness and its sensitivity to random variation.
This practice is well-established in many communities (e.g., \gls{abk:ml}), where it is considered a standard for demonstrating that the reported performance is stable and generalizable~\cite{ahmed2022managing}.

\subsection{Reproducibility of code}
In addition to the methodological transparency discussed throughout this paper, code availability and clarity are essential to the reproducibility of \gls{abk:amod} research. If the code is publicly available, reproducibility of code entails the ability to regenerate published results using the original code base and documentation, ideally without requiring deep knowledge of the underlying implementation. To achieve this, the code repository should contain a clear entry point (e.g., a README file), clearly specify data sources (or describe metadata if the data cannot be shared), provide environment setup instructions (e.g., requirements files or Docker containers), and include executable examples of how to run the experiments. Furthermore, to make future extensions or adaptations to new problem settings more accessible, it is recommended that authors modularize their code base, follow established style conventions, and provide clear inline documentation.

On the other hand, when code is not publicly accessible, reproducibility refers to the ability of independent researchers to re-implement and validate the proposed method or framework using the materials provided in the original publication, such as the paper, appendices, and supplementary documentation. In this case, it is crucial that these materials include unambiguous explanations of all algorithmic steps (e.g., well-organized and clear pseudocode), an account of any non-trivial engineering decisions or system-level dependencies (e.g., references to external libraries or solvers used in the original implementation), and sufficient information to reconstruct the experimental setup (e.g., hyperparameter configurations and data preprocessing steps). 
\rev{These elements help ensure that other researchers can replicate the original work without needing to reverse-engineer all implementation details.}

Recent efforts in the \gls{abk:ml} and systems research communities have underscored the importance of reproducible code and have led to stronger norms around sharing open-source software.
For guidance, readers may refer to~\cite{wu2024reproducibility}, which provides a hands-on tutorial for improving both reproducibility and replicability in control and learning pipelines.
A strong example of reproducibility best practices in the \gls{abk:amod} domain is demonstrated in~\cite{stanfordasl_rl4amod_2024}, where the code repository is well-documented, modularly organized, and aligned with many of the guidelines outlined \rev{in this work}.

\subsection{Corpus-level Reproducibility Assessment}
\begin{figure}
    \centering
    \includegraphics[width=0.8\linewidth]{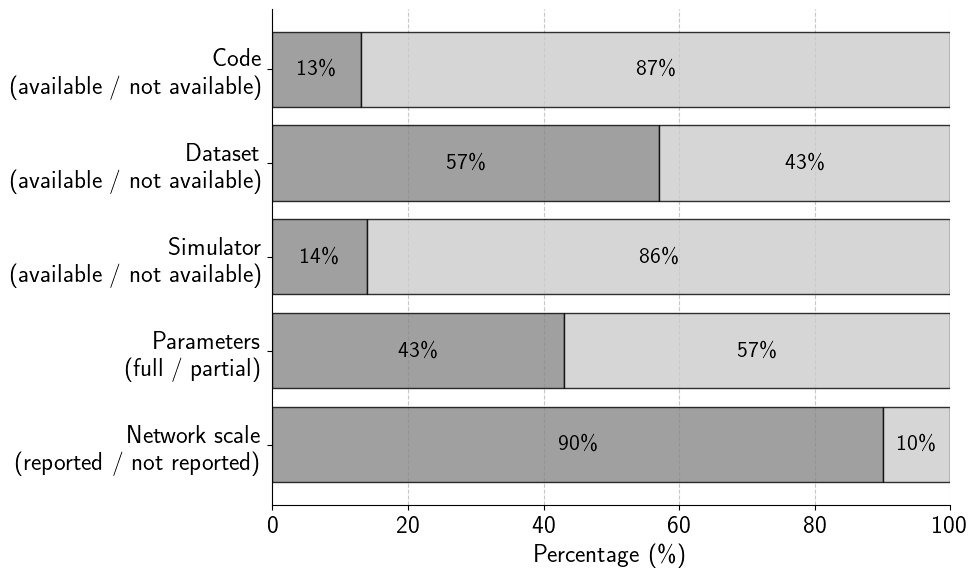}
    \caption{\rev{Statistics on reproducibility metrics.}}
    \label{fig:statistics}
\end{figure}

\rev{We conduct a rubric-based, corpus-level audit of peer-reviewed \gls{abk:amod} control papers (2015–2025) to quantify current reproducibility practices. 
The rubric covers five dimensions aligned with our checklist: \emph{code availability}, \emph{dataset accessibility}, \emph{simulator transparency}, \emph{parameter disclosure}, and \emph{reported network scale} (size/granularity).
\Cref{fig:statistics} summarizes aggregate proportions across the corpus, while \Cref{tab:analysis} presents an excerpted per-paper summary for ten highly cited works in robotics venues.\footnote{\rev{Notes for \cref{tab:analysis}: \checkmark indicates availability for code, dataset and simulator, fully reported for parameters, and real-world scale for network; \ding{53} indicates missing for code, dataset and simulator, partially reported for parameters, and small scale or not reported for network.}}{}
The main observations are:}

\rev{\noindent \textbf{1) Limited code availability.} The scarcity of accessible, runnable code is a primary reproducibility bottleneck: without artifacts, independent verification, extension, and fair benchmarking are very difficult even when methodology and results are clearly described.}

\rev{\noindent \textbf{2) Documentation gaps in released code.} Among the few studies that share repositories, $\approx$40\% lack basic run instructions or an environment lockfile. Open-source releases should minimally include a README with one-command execution, fixed seeds, and a pinned environment (e.g., Docker/Conda).}

\rev{\noindent \textbf{3) Underdocumented preprocessing.} Although over half of the papers use public datasets, many apply nontrivial preprocessing (e.g., demand re-partitioning or source merges) without releasing scripts or processed artifacts, making faithful reconstruction costly. Preprocessing code and checksums should be provided.}

\rev{\noindent \textbf{4) Small-scale evaluations.} Despite reporting network scale, about 36\% evaluate only on synthetic or heavily aggregated networks, limiting transfer to real-world city scale. Such results should be explicitly labeled as small-scale, with scalability discussed or tested across multiple granularities.}

\begin{table}[t]
\centering
\small
\resizebox{\columnwidth}{!}{
\begin{tabular}{|c|c|c|c|c|c|c|}
\hline
\textbf{Reference} & \textbf{Venue} & \textbf{Code} & \textbf{Dataset} & \textbf{Simulator} & \textbf{Parameters} & \textbf{Network} \\ \hline
\cite{pavone2012robotic} & IJRR & \ding{53} & \ding{53} & \ding{53} & \checkmark & \ding{53} \\ \hline
\cite{zhang2016control} & IJRR & \ding{53} & \checkmark & \ding{53} & \ding{53} & \checkmark \\ \hline
\cite{zhang2016model} & ICRA & \ding{53} & \checkmark & \ding{53} & \ding{53} & \ding{53} \\ \hline
\cite{rossi2018routing} & \shortstack{Springer AR} & \ding{53} & \ding{53} & \checkmark & \ding{53} & \checkmark \\ \hline
\cite{iglesias2018data} & ICRA & \ding{53} & \ding{53} & \checkmark & \checkmark & \checkmark \\ \hline
\cite{alonso2017predictive} & IROS & \ding{53} & \checkmark & \ding{53} & \ding{53} & \ding{53} \\ \hline
\cite{wallar2018vehicle} & IROS & \ding{53} & \checkmark & \ding{53} & \checkmark & \checkmark \\ \hline
\cite{iglesias2019bcmp} & IJRR & \ding{53} & \checkmark & \ding{53} & \checkmark & \checkmark \\ \hline
\cite{tsao2019model} & ICRA & \ding{53} & \checkmark & \ding{53} & \checkmark & \checkmark \\ \hline
\cite{miller2017predictive} & ICRA & \ding{53} & \ding{53} & \ding{53} & \ding{53} & \ding{53} \\ \hline
\end{tabular}
}
\caption{\rev{Reproducibility of a sample of studies.}}
\label{tab:analysis}
\end{table}

\rev{
Complementary to the aggregate statistics, the review highlights four issues that are \emph{not} about general code availability. 
First, \textbf{simulator accessibility}: there is no de facto standard, and even when custom or general-purpose tools are used, configuration/scenario files, random seeds, and state-transition logic are often not released, hampering exact replication. 
Second, \textbf{reproducible preprocessing}: many studies rely on well-known public datasets yet apply undocumented steps (e.g., demand re-partitioning or source merges; cf.~\cref{fig:scenario}) without publishing scripts, checksums, or processed artifacts. 
Third, \textbf{completeness of parameterization}: submodule parameters (e.g., energy models for electric fleets, pricing/cost components, solver tolerances/hyperparameters) are frequently omitted, making faithful reimplementation difficult. 
Fourth, \textbf{claims–evidence alignment}: scalability or efficiency claims are sometimes evaluated on limited scales or with non-comparable splits/baselines, weakening external validity. 
These gaps motivate our checklist items on releasing simulator configs, data-processing scripts or artifacts, and full parameter tables (including submodules), as well as seeded, variance-reported experiments on shared splits.
}

%% file: chapters/interaction.tex
\section{Interactions} \label{sec:interaction}
\rev{In multimodal transportation systems, \gls{abk:amod} services interact with public transit~\cite{zardini2019towards,vakayil2017integrating}, power grids~\cite{rossi2019interaction,estandia2021interaction}, other transport modes~\cite{zardini2023strategic}, and competing or cooperating mobility providers~\cite{9564501, he2025hierarchicalstrategicdecisionmakinglayered}. 
Explicitly modeling these couplings is essential to assess the systemic impact of \gls{abk:amod} deployments and their integration into existing urban infrastructure.}
When studying such interactions, the modeling, again, should be explicitly formalized within the methodology, as it significantly influences the realism and interpretability of the analysis.

\rev{A major research thread is the integration of \gls{abk:amod} with public transit. 
This is typically modeled either as a first-/last-mile connector to existing transit~\cite{ma2019dynamic} or as a fully intermodal system in which \gls{abk:amod} and transit are jointly optimized for end-to-end mobility~\cite{salazar2018interaction,salazar2019intermodal,zardini2019towards,vakayil2017integrating}. 
Early work relied largely on simulation~\cite{bischoff2017re}; more recent approaches introduce formal intermodal network models, where \gls{abk:amod} and transit form separate layers of a time-expanded graph with interlayer arcs encoding transfers~\cite{salazar2018interaction,salazar2019intermodal}. 
This structure supports non-myopic, coordinated control via \gls{abk:mpc}.}

\rev{Moving beyond control under fixed infrastructure, co-design frameworks jointly optimize \gls{abk:amod} and public transit~\cite{zardini2022co,Zardini2023}. 
Rather than posing a single monolithic problem, these methods decompose the system into interacting subproblems coupled through shared constraints (e.g., demand or resource limits), enabling scalable and modular design. 
This compositional view is particularly relevant for future multimodal systems, where \gls{abk:amod} must be integrated with, rather than simply replace, other modes to avoid cannibalizing public transit ridership~\cite{oh2020assessing}. 
Co-design principles from category theory offer a pathway toward general, reusable architectures for multimodal mobility~\cite{zardini2022co,zardini2020co,zardini2021co,zardini2021co_hardware}.}

\rev{A second key interaction concerns electric \gls{abk:amod} fleets and power grids~\cite{rossi2019interaction,estandia2021interaction}. \glspl{abk:ev} act simultaneously as mobility resources and electrical loads: large-scale charging can stress local distribution networks and affect grid stability, while tariffs and demand-side management policies shape routing and charging decisions. 
Grid-aware control, through price-responsive dispatch, load balancing, or rebalancing strategies that account for grid constraints, offers a promising avenue to jointly ensure transportation efficiency and power system reliability under increasing electrification.}

\rev{Finally, interactions across modes are mediated by passenger \emph{mode choice}. 
In multimodal networks, users select among \gls{abk:amod}, transit, and other options based on sociodemographics and trip-specific attributes such as time and cost. These behaviors are typically modeled via discrete choice models~\cite{steck2018autonomous,liu2017tracking}. Incorporating such models into \gls{abk:amod} control is crucial for realistically capturing substitution and complementarity between modes and for assessing the real-world performance of control policies~\cite{kamel2019exploring}.}

%% file: chapters/conclusion.tex
\section{Conclusion: Toward a culture of reproducible \glsentrytext{abk:amod} research} \label{sec:conclusion}
\rev{\gls{abk:amod} systems sit at the intersection of robotics, control, transportation, and optimization, and operate ``in the wild,'' where demand, congestion, failures, and human–infrastructure interactions are nonstationary. In this setting, rigor and traceability are not optional—they are prerequisites for credible scientific progress. 
This survey has presented a system-level view of reproducibility in \emph{fleet} robotics, with \gls{abk:amod} as a primary use case, and distilled it into a practical playbook: runnable artifacts with pinned environments, published simulator configurations and scenario files, scripted preprocessing or released processed splits, fixed seeds with variance reporting, complete parameter tables, shared splits and baselines, and explicit reporting of scale and abstraction (node/mixed/region). 
These practices enable fair comparison, sim-to-real traceability, and defensible safety arguments across ground, aerial, marine, and service-robot deployments, while standardized terminology reduces fragmentation between optimization- and learning-based approaches.}

\rev{To move toward a more reproducible research culture, we recommend that every \gls{abk:amod} paper clearly specify:}

\begin{enumerate}
    \item \textbf{The system model} used to represent \gls{abk:amod}.
    \item \textbf{The specific operation and related constraints}. Make sure to be concise with the terminologies when describing the operations by sticking to the definitions provided in Section \ref{sec:operation}.
    \item \textbf{The objective} of the control problem.
    \item \textbf{Assumptions} regarding the demand, congestion, passenger model, and uncertainty.
    \item \textbf{The algorithm} used to derive the solution.
    \item \textbf{Detailed experiment setup}, which should include:
    \begin{itemize}
        \item Description and preprocessing steps of the utilized dataset, along with a link to the source.
        \item The simulation environment.
        \item The level of network granularity.
        \item Simulation and algorithm parameters, including the rationale for specific parameter values, if applicable.
        \item Evaluation criteria for the algorithm, which should align with the stated contributions of the research.
    \end{itemize}
\end{enumerate}

Beyond the paper, \textbf{code} is another important deliverable for \gls{abk:amod} research. However, sharing code has not been a norm in transportation research~\cite{riehl2025revisiting}. To ensure reproducibility, the following guidelines should be followed for code:
\begin{enumerate}
    \item The code should be publicly available.
    \item It should include a detailed README with step-by-step instructions for setting up the environment and running an instance.
    \item The code should be modularized, well-documented, and adhere to style conventions.
\end{enumerate}

\begin{figure}
    \centering
    \includegraphics[width=1.0\linewidth]{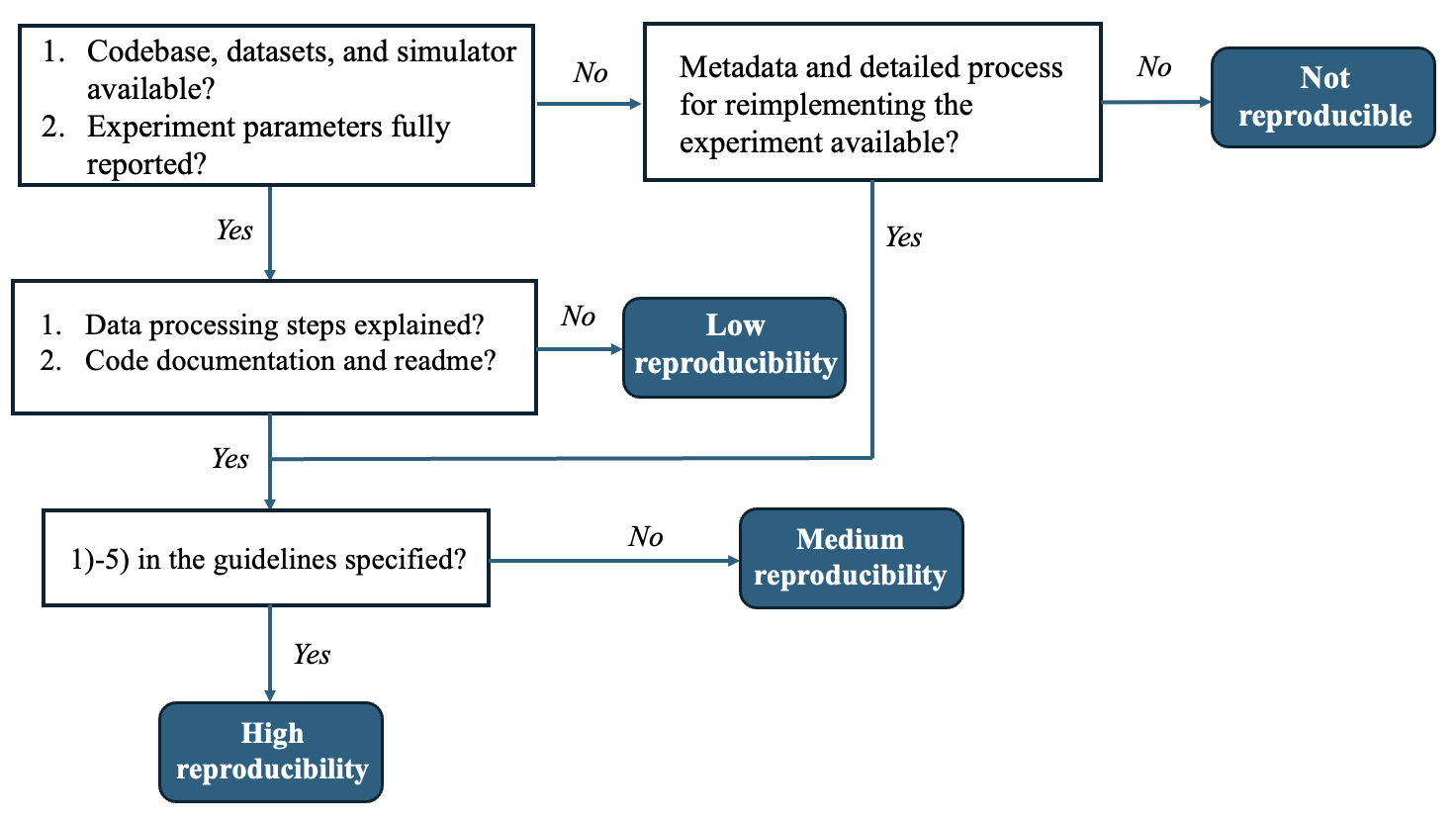}
    \caption{\rev{Tiered scoring procedure for reproducibility evaluation.}}
    \label{fig:score}
\end{figure}

\rev{Alongside these guidelines, we provide a tiered scoring procedure (\cref{fig:score}) to help authors self-assess the reproducibility level of their work and to give reviewers and readers a simple, transparent rubric for evaluation.}

\rev{Looking ahead, emerging trends in artificial intelligence, particularly large language models (LLMs) and agentic AI systems, will further reshape how reproducibility is addressed.
These tools can strengthen reproducibility by reducing the overhead of creating high-quality artifacts, for example via automated extraction of assumptions and configuration details from code, generation of standardized documentation, and translation or refactoring of legacy implementations.
Agentic systems can also help automate test generation and systematic design-space exploration.
At the same time, AI-assisted workflows introduce new pitfalls: they may rely on opaque intermediate steps, stochastic outputs, or evolving externally hosted models, making it harder to reconstruct how results were produced.
In robotics, where complex software stacks interface with simulation or hardware, such hidden variability can seriously undermine reproducibility if undocumented.
AI tools should therefore be treated as explicit, well-specified components of the pipeline, with clear disclosure of model versions, prompts or policies, and agent behaviors.}

Overall, our central message is clear: \textbf{reproducibility is not a peripheral concern, it is a foundational principle of rigorous science}. 
Without it, innovations are difficult to verify, compare, or build upon.
With it, we can create a research ecosystem that is transparent, collaborative, and scalable.

Although our focus is on \gls{abk:amod}, the principles we advocate extend naturally to other domains involving networked autonomy, multimodal mobility, and data-driven control. 
We encourage researchers, reviewers, and publishers to treat reproducibility as a shared responsibility and to embed it into the design of experiments, not merely as a post-hoc concern.

%% file: cas-refs.bib
@book{larson1981urban,
  title={Urban operations research},
  author={Larson, Richard C and Odoni, Amedeo R},
  publisher={Prentice-Hall},
  year={1981}
}

@phdthesis{Zardini2023,
  author = {Zardini, Gioele},
  title = {Co-Design of Complex Systems: From Autonomy to Future Mobility Systems},
  year = {2023},
  url = {https://www.research-collection.ethz.ch/handle/20.500.11850/648075},
  school = {ETH Zürich, Switzerland},
}

@article{kim2024estimate,
  title={Estimate Then Predict: Convex Formulation for Travel Demand Forecasting},
  author={Kim, Youngseo and Zardini, Gioele and Samaranayake, Samitha and Shafiee, Soroosh},
  journal={Available at SSRN 4977199},
  year={2024}
}

@article{zhang2016control,
  title={Control of robotic mobility-on-demand systems: a queueing-theoretical perspective},
  author={Zhang, Rick and Pavone, Marco},
  journal={The International Journal of Robotics Research},
  volume={35},
  number={1-3},
  pages={186--203},
  year={2016},
  publisher={SAGE Publications Sage UK: London, England}
}

@article{iglesias2019bcmp,
  title={A BCMP network approach to modeling and controlling autonomous mobility-on-demand systems},
  author={Iglesias, Ramon and Rossi, Federico and Zhang, Rick and Pavone, Marco},
  journal={The International Journal of Robotics Research},
  volume={38},
  number={2-3},
  pages={357--374},
  year={2019},
  publisher={SAGE Publications Sage UK: London, England}
}

@article{zhang2018analysis,
  title={Analysis, control, and evaluation of mobility-on-demand systems: A queueing-theoretical approach},
  author={Zhang, Rick and Rossi, Federico and Pavone, Marco},
  journal={IEEE Transactions on Control of Network Systems},
  volume={6},
  number={1},
  pages={115--126},
  year={2018},
  publisher={IEEE}
}

@inproceedings{spieser2016shared,
  title={Shared-vehicle mobility-on-demand systems: A fleet operator’s guide to rebalancing empty vehicles},
  author={Spieser, Kevin and Samaranayake, Samitha and Gruel, Wolfgang and Frazzoli, Emilio},
  booktitle={Transportation Research Board 95th Annual Meeting},
  number={16-5987},
  year={2016},
  organization={Transportation Research Board}
}

@article{pavone2012robotic,
  title={Robotic load balancing for mobility-on-demand systems},
  author={Pavone, Marco and Smith, Stephen L and Frazzoli, Emilio and Rus, Daniela},
  journal={The International Journal of Robotics Research},
  volume={31},
  number={7},
  pages={839--854},
  year={2012},
  publisher={Sage Publications Sage UK: London, England}
}

@article{wollenstein2021routing,
  title={Routing and rebalancing intermodal autonomous mobility-on-demand systems in mixed traffic},
  author={Wollenstein-Betech, Salom{\'o}n and Salazar, Mauro and Houshmand, Arian and Pavone, Marco and Paschalidis, Ioannis Ch and Cassandras, Christos G},
  journal={IEEE Transactions on Intelligent Transportation Systems},
  volume={23},
  number={8},
  pages={12263--12275},
  year={2021},
  publisher={IEEE}
}

@inproceedings{enders2023hybrid,
  title={Hybrid multi-agent deep reinforcement learning for autonomous mobility on demand systems},
  author={Enders, Tobias and Harrison, James and Pavone, Marco and Schiffer, Maximilian},
  booktitle={Learning for Dynamics and Control Conference},
  pages={1284--1296},
  year={2023},
  organization={PMLR}
}

@article{woywood2024multi,
  title={Multi-Agent Soft Actor-Critic with Global Loss for Autonomous Mobility-on-Demand Fleet Control},
  author={Woywood, Zeno and Wiltfang, Jasper I and Luy, Julius and Enders, Tobias and Schiffer, Maximilian},
  journal={arXiv preprint arXiv:2404.06975},
  year={2024}
}

@article{hu2020artificial,
  title={An artificial-neural-network-based model for real-time dispatching of electric autonomous taxis},
  author={Hu, Liang and Dong, Jing},
  journal={IEEE Transactions on Intelligent Transportation Systems},
  volume={23},
  number={2},
  pages={1519--1528},
  year={2020},
  publisher={IEEE}
}

@article{lazar2021learning,
  title={Learning how to dynamically route autonomous vehicles on shared roads},
  author={Lazar, Daniel A and B{\i}y{\i}k, Erdem and Sadigh, Dorsa and Pedarsani, Ramtin},
  journal={Transportation research part C: emerging technologies},
  volume={130},
  pages={103258},
  year={2021},
  publisher={Elsevier}
}

@inproceedings{bei2018algorithms,
  title={Algorithms for trip-vehicle assignment in ride-sharing},
  author={Bei, Xiaohui and Zhang, Shengyu},
  booktitle={Proceedings of the AAAI Conference on Artificial Intelligence},
  volume={32},
  number={1},
  year={2018}
}

@article{narayanan2020shared,
  title={Shared autonomous vehicle services: A comprehensive review},
  author={Narayanan, Santhanakrishnan and Chaniotakis, Emmanouil and Antoniou, Constantinos},
  journal={Transportation Research Part C: Emerging Technologies},
  volume={111},
  pages={255--293},
  year={2020},
  publisher={Elsevier}
}

@inproceedings{chen2017hierarchical,
  title={Hierarchical data-driven vehicle dispatch and ride-sharing},
  author={Chen, Ximing and Miao, Fei and Pappas, George J and Preciado, Victor},
  booktitle={2017 IEEE 56th Annual Conference on Decision and Control (CDC)},
  pages={4458--4463},
  year={2017},
  organization={IEEE}
}

@article{levin2017congestion,
  title={Congestion-aware system optimal route choice for shared autonomous vehicles},
  author={Levin, Michael W},
  journal={Transportation Research Part C: Emerging Technologies},
  volume={82},
  pages={229--247},
  year={2017},
  publisher={Elsevier}
}

@article{kang2021maximum,
  title={Maximum-stability dispatch policy for shared autonomous vehicles},
  author={Kang, Di and Levin, Michael W},
  journal={Transportation Research Part B: Methodological},
  volume={148},
  pages={132--151},
  year={2021},
  publisher={Elsevier}
}

@article{dandl2019evaluating,
  title={Evaluating the impact of spatio-temporal demand forecast aggregation on the operational performance of shared autonomous mobility fleets},
  author={Dandl, Florian and Hyland, Michael and Bogenberger, Klaus and Mahmassani, Hani S},
  journal={Transportation},
  volume={46},
  number={6},
  pages={1975--1996},
  year={2019},
  publisher={Springer}
}

@article{fagnant2018dynamic,
  title={Dynamic ride-sharing and fleet sizing for a system of shared autonomous vehicles in Austin, Texas},
  author={Fagnant, Daniel J and Kockelman, Kara M},
  journal={Transportation},
  volume={45},
  pages={143--158},
  year={2018},
  publisher={Springer}
}

@article{yang2020planning,
  title={Planning and operations of mixed fleets in mobility-on-demand systems},
  author={Yang, Kaidi and Tsao, Matthew W and Xu, Xin and Pavone, Marco},
  journal={arXiv preprint arXiv:2008.08131},
  year={2020}
}

@article{vazifeh2018addressing,
  title={Addressing the minimum fleet problem in on-demand urban mobility},
  author={Vazifeh, Mohammad M and Santi, Paolo and Resta, Giovanni and Strogatz, Steven H and Ratti, Carlo},
  journal={Nature},
  volume={557},
  number={7706},
  pages={534--538},
  year={2018},
  publisher={Nature Publishing Group}
}

@article{ma2017designing,
  title={Designing optimal autonomous vehicle sharing and reservation systems: A linear programming approach},
  author={Ma, Jiaqi and Li, Xiaopeng and Zhou, Fang and Hao, Wei},
  journal={Transportation Research Part C: Emerging Technologies},
  volume={84},
  pages={124--141},
  year={2017},
  publisher={Elsevier}
}

@inproceedings{volkov2012markov,
  title={Markov-based redistribution policy model for future urban mobility networks},
  author={Volkov, Mikhail and Aslam, Javed and Rus, Daniela},
  booktitle={2012 15th International IEEE Conference on Intelligent Transportation Systems},
  pages={1906--1911},
  year={2012},
  organization={IEEE}
}

@inproceedings{li2019efficient,
  title={An efficient matching method for dispatching autonomous vehicles},
  author={Li, Ming and Zheng, Nan and Wu, Xinkai and Huo, Xiang},
  booktitle={2019 IEEE Intelligent Transportation Systems Conference (ITSC)},
  pages={3013--3018},
  year={2019},
  organization={IEEE}
}

@inproceedings{spieser2016vehicle,
  title={Vehicle routing for shared-mobility systems with time-varying demand},
  author={Spieser, Kevin and Samaranayake, Samitha and Frazzoli, Emilio},
  booktitle={2016 American Control Conference (ACC)},
  pages={796--802},
  year={2016},
  organization={IEEE}
}

@inproceedings{boewing2020vehicle,
  title={A vehicle coordination and charge scheduling algorithm for electric autonomous mobility-on-demand systems},
  author={Boewing, Felix and Schiffer, Maximilian and Salazar, Mauro and Pavone, Marco},
  booktitle={2020 American Control Conference (ACC)},
  pages={248--255},
  year={2020},
  organization={IEEE}
}

@article{belakaria2019multi,
  title={Multi-class management with sub-class service for autonomous electric mobility on-demand systems},
  author={Belakaria, Syrine and Ammous, Mustafa and Smith, Lauren and Sorour, Sameh and Abdel-Rahim, Ahmed},
  journal={IEEE Transactions on Vehicular Technology},
  volume={68},
  number={7},
  pages={7155--7159},
  year={2019},
  publisher={IEEE}
}

@article{guo2020vehicle,
  title={Vehicle rebalancing with charging scheduling in one-way car-sharing systems},
  author={Guo, Ge and Xu, Tao},
  journal={IEEE Transactions on Intelligent Transportation Systems},
  volume={23},
  number={5},
  pages={4342--4351},
  year={2020},
  publisher={IEEE}
}

@article{iacobucci2018modeling,
  title={Modeling shared autonomous electric vehicles: Potential for transport and power grid integration},
  author={Iacobucci, Riccardo and McLellan, Benjamin and Tezuka, Tetsuo},
  journal={Energy},
  volume={158},
  pages={148--163},
  year={2018},
  publisher={Elsevier}
}

@article{iacobucci2019optimization,
  title={Optimization of shared autonomous electric vehicles operations with charge scheduling and vehicle-to-grid},
  author={Iacobucci, Riccardo and McLellan, Benjamin and Tezuka, Tetsuo},
  journal={Transportation Research Part C: Emerging Technologies},
  volume={100},
  pages={34--52},
  year={2019},
  publisher={Elsevier}
}

@article{chu2021joint,
  title={Joint rebalancing and vehicle-to-grid coordination for autonomous vehicle public transportation system},
  author={Chu, Kai-Fung and Lam, Albert YS and Li, Victor OK},
  journal={IEEE Transactions on Intelligent Transportation Systems},
  volume={23},
  number={7},
  pages={7156--7169},
  year={2021},
  publisher={IEEE}
}

@article{nair2011fleet,
  title={Fleet management for vehicle sharing operations},
  author={Nair, Rahul and Miller-Hooks, Elise},
  journal={Transportation Science},
  volume={45},
  number={4},
  pages={524--540},
  year={2011},
  publisher={INFORMS}
}

@article{duan2020centralized,
  title={Centralized and decentralized autonomous dispatching strategy for dynamic autonomous taxi operation in hybrid request mode},
  author={Duan, Leyi and Wei, Yuguang and Zhang, Jinchuan and Xia, Yang},
  journal={Transportation Research Part C: Emerging Technologies},
  volume={111},
  pages={397--420},
  year={2020},
  publisher={Elsevier}
}

@article{hyland2018dynamic,
  title={Dynamic autonomous vehicle fleet operations: Optimization-based strategies to assign AVs to immediate traveler demand requests},
  author={Hyland, Michael and Mahmassani, Hani S},
  journal={Transportation Research Part C: Emerging Technologies},
  volume={92},
  pages={278--297},
  year={2018},
  publisher={Elsevier}
}

@article{huang2024bi,
  title={A bi-level approach for optimal vehicle relocating in Mobility-On-Demand systems with approximate dynamic programming and coverage control},
  author={Huang, Yunping and Zhu, Pengbo and Zhong, Renxin and Geroliminis, Nikolas},
  journal={Transportation Research Part E: Logistics and Transportation Review},
  volume={192},
  pages={103754},
  year={2024},
  publisher={Elsevier}
}

@article{neuburger1971economics,
  title={Economics of heavily congested roads},
  author={Neuburger, Henry},
  journal={Transportation Research/UK/},
  year={1971}
}

@Inbook{Pavone2015,
author="Pavone, Marco",
editor="Maurer, Markus
and Gerdes, J. Christian
and Lenz, Barbara
and Winner, Hermann",
title="Autonomous Mobility-on-Demand Systems for Future Urban Mobility",
bookTitle="Autonomes Fahren: Technische, rechtliche und gesellschaftliche Aspekte",
year="2015",
publisher="Springer Berlin Heidelberg",
address="Berlin, Heidelberg",
pages="399--416",
isbn="978-3-662-45854-9",
doi="10.1007/978-3-662-45854-9_19",
url="https://doi.org/10.1007/978-3-662-45854-9_19"
}

@inproceedings{miao2017data,
  title={Data-driven distributionally robust vehicle balancing using dynamic region partitions},
  author={Miao, Fei and Han, Shuo and Hendawi, Abdeltawab M and Khalefa, Mohamed E and Stankovic, John A and Pappas, George J},
  booktitle={Proceedings of the 8th International Conference on Cyber-Physical Systems},
  pages={261--271},
  year={2017}
}

@inproceedings{he2023robust,
  title={A Robust and Constrained Multi-Agent Reinforcement Learning Electric Vehicle Rebalancing Method in AMoD Systems},
  author={He, Sihong and Wang, Yue and Han, Shuo and Zou, Shaofeng and Miao, Fei},
  booktitle={2023 IEEE/RSJ International Conference on Intelligent Robots and Systems (IROS)},
  pages={5637--5644},
  year={2023},
  organization={IEEE}
}

@article{guo2021robust,
  title={Robust matching-integrated vehicle rebalancing in ride-hailing system with uncertain demand},
  author={Guo, Xiaotong and Caros, Nicholas S and Zhao, Jinhua},
  journal={Transportation Research Part B: Methodological},
  volume={150},
  pages={161--189},
  year={2021},
  publisher={Elsevier}
}

@article{bischoff2016simulation,
  title={Simulation of city-wide replacement of private cars with autonomous taxis in Berlin},
  author={Bischoff, Joschka and Maciejewski, Michal},
  journal={Procedia computer science},
  volume={83},
  pages={237--244},
  year={2016},
  publisher={Elsevier}
}

@inproceedings{zhang2016model,
  title={Model predictive control of autonomous mobility-on-demand systems},
  author={Zhang, Rick and Rossi, Federico and Pavone, Marco},
  booktitle={2016 IEEE international conference on robotics and automation (ICRA)},
  pages={1382--1389},
  year={2016},
  organization={IEEE}
}

@article{sayarshad2017non,
  title={Non-myopic relocation of idle mobility-on-demand vehicles as a dynamic location-allocation-queueing problem},
  author={Sayarshad, Hamid R and Chow, Joseph YJ},
  journal={Transportation Research Part E: Logistics and Transportation Review},
  volume={106},
  pages={60--77},
  year={2017},
  publisher={Elsevier}
}

@inproceedings{lin2018efficient,
  title={Efficient large-scale fleet management via multi-agent deep reinforcement learning},
  author={Lin, Kaixiang and Zhao, Renyu and Xu, Zhe and Zhou, Jiayu},
  booktitle={Proceedings of the 24th ACM SIGKDD international conference on knowledge discovery \& data mining},
  pages={1774--1783},
  year={2018}
}

@inproceedings{han2016routing,
  title={Routing an autonomous taxi with reinforcement learning},
  author={Han, Miyoung and Senellart, Pierre and Bressan, St{\'e}phane and Wu, Huayu},
  booktitle={Proceedings of the 25th ACM International on Conference on Information and Knowledge Management},
  pages={2421--2424},
  year={2016}
}

@inproceedings{jindal2018optimizing,
  title={Optimizing taxi carpool policies via reinforcement learning and spatio-temporal mining},
  author={Jindal, Ishan and Qin, Zhiwei Tony and Chen, Xuewen and Nokleby, Matthew and Ye, Jieping},
  booktitle={2018 IEEE International Conference on Big Data (Big Data)},
  pages={1417--1426},
  year={2018},
  organization={IEEE}
}

@article{rossi2018routing,
  title={Routing autonomous vehicles in congested transportation networks: Structural properties and coordination algorithms},
  author={Rossi, Federico and Zhang, Rick and Hindy, Yousef and Pavone, Marco},
  journal={Autonomous Robots},
  volume={42},
  pages={1427--1442},
  year={2018},
  publisher={Springer}
}

@article{albert2019imbalance,
  title={Imbalance in mobility-on-demand systems: A stochastic model and distributed control approach},
  author={Albert, Marc and Ruch, Claudio and Frazzoli, Emilio},
  journal={ACM Transactions on Spatial Algorithms and Systems (TSAS)},
  volume={5},
  number={2},
  pages={1--22},
  year={2019},
  publisher={ACM New York, NY, USA}
}

@article{levin2017general,
  title={A general framework for modeling shared autonomous vehicles with dynamic network-loading and dynamic ride-sharing application},
  author={Levin, Michael W and Kockelman, Kara M and Boyles, Stephen D and Li, Tianxin},
  journal={Computers, Environment and Urban Systems},
  volume={64},
  pages={373--383},
  year={2017},
  publisher={Elsevier}
}

@inproceedings{gachter2021image,
  title={Image representation of a city and its taxi fleet for end-to-end learning of rebalancing policies},
  author={G{\"a}chter, Joel and Zanardi, Alessandro and Ruch, Claudio and Frazzoli, Emilio},
  booktitle={2021 IEEE International Conference on Robotics and Automation (ICRA)},
  pages={8076--8082},
  year={2021},
  organization={IEEE}
}

@article{jungel2023learning,
  title={Learning-based online optimization for autonomous mobility-on-demand fleet control},
  author={Jungel, Kai and Parmentier, Axel and Schiffer, Maximilian and Vidal, Thibaut},
  journal={arXiv preprint arXiv:2302.03963},
  year={2023}
}

@misc{wu2024reproducibility,
  title = {Reproducibility in Transportation Research: A Hands-on Tutorial},
  author = {Wu, Cathy and Ghosh, Bidisha and Zheng, Zuduo and Mart{\'i}nez, Irene},
  year = {2024},
  publisher = {IEEE International Conference on Intelligent Transportation Systems (ITSC)},
  url = {https://www.rerite.org/itsc24-rr-tutorial/}
}

@article{salazar2019intermodal,
  title={Intermodal autonomous mobility-on-demand},
  author={Salazar, Mauro and Lanzetti, Nicolas and Rossi, Federico and Schiffer, Maximilian and Pavone, Marco},
  journal={IEEE Transactions on Intelligent Transportation Systems},
  volume={21},
  number={9},
  pages={3946--3960},
  year={2019},
  publisher={IEEE}
}

@inproceedings{salazar2018interaction,
  title={On the interaction between autonomous mobility-on-demand and public transportation systems},
  author={Salazar, Mauro and Rossi, Federico and Schiffer, Maximilian and Onder, Christopher H and Pavone, Marco},
  booktitle={2018 21st international conference on intelligent transportation systems (ITSC)},
  pages={2262--2269},
  year={2018},
  organization={IEEE}
}

@inproceedings{zgraggen2019model,
  title={A model predictive control scheme for intermodal autonomous mobility-on-demand},
  author={Zgraggen, Jannik and Tsao, Matthew and Salazar, Mauro and Schiffer, Maximilian and Pavone, Marco},
  booktitle={2019 IEEE Intelligent Transportation Systems Conference (ITSC)},
  pages={1953--1960},
  year={2019},
  organization={IEEE}
}

@article{zardini2019towards,
  title={Towards a co-design framework for future mobility systems},
  author={Zardini, Gioele and Lanzetti, Nicolas and Salazar, Mauro and Censi, Andrea and Frazzoli, Emilio and Pavone, Marco},
  journal={arXiv preprint arXiv:1910.07714},
  year={2019}
}

@article{ma2019dynamic,
  title={A dynamic ridesharing dispatch and idle vehicle repositioning strategy with integrated transit transfers},
  author={Ma, Tai-Yu and Rasulkhani, Saeid and Chow, Joseph YJ and Klein, Sylvain},
  journal={Transportation Research Part E: Logistics and Transportation Review},
  volume={128},
  pages={417--442},
  year={2019},
  publisher={Elsevier}
}

@inproceedings{bischoff2017re,
  title={Re-defining the role of public transport in a world of shared autonomous vehicles},
  author={Bischoff, Joschka and Kaddoura, Ihab and Maciejewski, Michal and Nagel, Kai},
  booktitle={Symposium of the European Association for Research in Transportation (hEART)},
  year={2017}
}

@techreport{vakayil2017integrating,
  title={Integrating shared-vehicle mobility-on-demand systems with public transit},
  author={Vakayil, Akhil and Gruel, Wolfgang and Samaranayake, Samitha},
  institution={Transportation Research Board},
  year={2017}
}

@article{rossi2019interaction,
  title={On the interaction between autonomous mobility-on-demand systems and the power network: Models and coordination algorithms},
  author={Rossi, Federico and Iglesias, Ramon and Alizadeh, Mahnoosh and Pavone, Marco},
  journal={IEEE Transactions on Control of Network Systems},
  volume={7},
  number={1},
  pages={384--397},
  year={2019},
  publisher={IEEE}
}

@article{estandia2021interaction,
  title={On the interaction between autonomous mobility on demand systems and power distribution networks—an optimal power flow approach},
  author={Estandia, Alvaro and Schiffer, Maximilian and Rossi, Federico and Luke, Justin and Kara, Emre Can and Rajagopal, Ram and Pavone, Marco},
  journal={IEEE Transactions on Control of Network Systems},
  volume={8},
  number={3},
  pages={1163--1176},
  year={2021},
  publisher={IEEE}
}

@book{shortle2018fundamentals,
  title={Fundamentals of queueing theory},
  author={Shortle, John F and Thompson, James M and Gross, Donald and Harris, Carl M},
  volume={399},
  year={2018},
  publisher={John Wiley \& Sons}
}

@inproceedings{hoppe2024global,
  title={Global rewards in multi-agent deep reinforcement learning for autonomous mobility on demand systems},
  author={Hoppe, Heiko and Enders, Tobias and Cappart, Quentin and Schiffer, Maximilian},
  booktitle={6th Annual Learning for Dynamics \& Control Conference},
  pages={260--272},
  year={2024},
  organization={PMLR}
}

@inproceedings{miller2017predictive,
  title={Predictive positioning and quality of service ridesharing for campus mobility on demand systems},
  author={Miller, Justin and How, Jonathan P},
  booktitle={2017 IEEE International Conference on Robotics and Automation (ICRA)},
  pages={1402--1408},
  year={2017},
  organization={IEEE}
}

@article{kim2024learning,
  title={Learning-Augmented Vehicle Dispatching with Slack Times for High-Capacity Ride-Pooling},
  author={Kim, Youngseo and Jayawardana, Vindula and Samaranayake, Samitha},
  journal={Available at SSRN 4801437},
  year={2024}
}

@inproceedings{shah2020neural,
  title={Neural approximate dynamic programming for on-demand ride-pooling},
  author={Shah, Sanket and Lowalekar, Meghna and Varakantham, Pradeep},
  booktitle={Proceedings of the AAAI Conference on Artificial Intelligence},
  volume={34},
  number={01},
  pages={507--515},
  year={2020}
}

@article{alonso2017demand,
  title={On-demand high-capacity ride-sharing via dynamic trip-vehicle assignment},
  author={Alonso-Mora, Javier and Samaranayake, Samitha and Wallar, Alex and Frazzoli, Emilio and Rus, Daniela},
  journal={Proceedings of the National Academy of Sciences},
  volume={114},
  number={3},
  pages={462--467},
  year={2017},
  publisher={National Acad Sciences}
}

@article{li2025learning,
  title={Learning Joint Rebalancing and Dynamic Pricing Policies for Autonomous Mobility-on-Demand},
  author={Li, Xinling and Schmidt, Carolin and Gammelli, Daniele and Rodrigues, Filipe},
  journal={IEEE Transactions on Intelligent Transportation Systems},
  year={2025},
  publisher={IEEE}
}

@misc{he2025hierarchicalstrategicdecisionmakinglayered,
      title={Hierarchical Strategic Decision-Making in Layered Mobility Systems}, 
      author={Mingjia He and Zhiyu He and Jan Ghadamian and Florian Dörfler and Emilio Frazzoli and Gioele Zardini},
      year={2025},
      eprint={2511.08734},
      archivePrefix={arXiv},
      primaryClass={eess.SY},
      url={https://arxiv.org/abs/2511.08734}, 
}

@article{li25speedup,
  author = {Li, Xinling and Gammelli, Daniele and Wallar, Alex and Zhao, Jinhua and Zardini, Gioele},
  title = {Accelerating High-Capacity Ridepooling in Robo-Taxi Systems},
  year = {2026},
  url = {https://zardini.mit.edu/assets/Li25-Speedup.pdf}
}

@article{schiffer2019vehicle,
  title={Vehicle routing and location routing with intermediate stops: A review},
  author={Schiffer, Maximilian and Schneider, Michael and Walther, Grit and Laporte, Gilbert},
  journal={Transportation Science},
  volume={53},
  number={2},
  pages={319--343},
  year={2019},
  publisher={INFORMS}
}

@inproceedings{li2021optimal,
  title={Optimal online dispatch for high-capacity shared autonomous mobility-on-demand systems},
  author={Li, Cheng and Parker, David and Hao, Qi},
  booktitle={2021 IEEE International Conference on Robotics and Automation (ICRA)},
  pages={779--785},
  year={2021},
  organization={IEEE}
}

@inproceedings{alonso2017predictive,
  title={Predictive routing for autonomous mobility-on-demand systems with ride-sharing},
  author={Alonso-Mora, Javier and Wallar, Alex and Rus, Daniela},
  booktitle={2017 IEEE/RSJ International Conference on Intelligent Robots and Systems (IROS)},
  pages={3583--3590},
  year={2017},
  organization={IEEE}
}

@article{fielbaum2021demand,
  title={On-demand ridesharing with optimized pick-up and drop-off walking locations},
  author={Fielbaum, Andres and Bai, Xiaoshan and Alonso-Mora, Javier},
  journal={Transportation research part C: emerging technologies},
  volume={126},
  pages={103061},
  year={2021},
  publisher={Elsevier}
}

@inproceedings{tsao2019model,
  title={Model predictive control of ride-sharing autonomous mobility-on-demand systems},
  author={Tsao, Matthew and Milojevic, Dejan and Ruch, Claudio and Salazar, Mauro and Frazzoli, Emilio and Pavone, Marco},
  booktitle={2019 International conference on robotics and automation (ICRA)},
  pages={6665--6671},
  year={2019},
  organization={IEEE}
}

@article{yu2019integrated,
  title={An integrated decomposition and approximate dynamic programming approach for on-demand ride pooling},
  author={Yu, Xian and Shen, Siqian},
  journal={IEEE Transactions on Intelligent Transportation Systems},
  volume={21},
  number={9},
  pages={3811--3820},
  year={2019},
  publisher={IEEE}
}

@article{simonetto2019real,
  title={Real-time city-scale ridesharing via linear assignment problems},
  author={Simonetto, Andrea and Monteil, Julien and Gambella, Claudio},
  journal={Transportation Research Part C: Emerging Technologies},
  volume={101},
  pages={208--232},
  year={2019},
  publisher={Elsevier}
}

@article{gueriau2020shared,
  title={Shared autonomous mobility on demand: A learning-based approach and its performance in the presence of traffic congestion},
  author={Gu{\'e}riau, Maxime and Cugurullo, Federico and Acheampong, Ransford A and Dusparic, Ivana},
  journal={IEEE Intelligent Transportation Systems Magazine},
  volume={12},
  number={4},
  pages={208--218},
  year={2020},
  publisher={IEEE}
}

@inproceedings{wallar2018vehicle,
  title={Vehicle rebalancing for mobility-on-demand systems with ride-sharing},
  author={Wallar, Alex and Van Der Zee, Menno and Alonso-Mora, Javier and Rus, Daniela},
  booktitle={2018 IEEE/RSJ international conference on intelligent robots and systems (IROS)},
  pages={4539--4546},
  year={2018},
  organization={IEEE}
}

@inproceedings{wen2017rebalancing,
  title={Rebalancing shared mobility-on-demand systems: A reinforcement learning approach},
  author={Wen, Jian and Zhao, Jinhua and Jaillet, Patrick},
  booktitle={2017 IEEE 20th international conference on intelligent transportation systems (ITSC)},
  pages={220--225},
  year={2017},
  organization={Ieee}
}

@article{tong2018unified,
  title={A unified approach to route planning for shared mobility},
  author={Tong, Yongxin and Zeng, Yuxiang and Zhou, Zimu and Chen, Lei and Ye, Jieping and Xu, Ke},
  journal={Proceedings of the VLDB Endowment},
  volume={11},
  number={11},
  pages={1633},
  year={2018},
  publisher={VLDB Endowment}
}

@inproceedings{gueriau2018samod,
  title={Samod: Shared autonomous mobility-on-demand using decentralized reinforcement learning},
  author={Gu{\'e}riau, Maxime and Dusparic, Ivana},
  booktitle={2018 21st International Conference on Intelligent Transportation Systems (ITSC)},
  pages={1558--1563},
  year={2018},
  organization={IEEE}
}

@article{al2019deeppool,
  title={Deeppool: Distributed model-free algorithm for ride-sharing using deep reinforcement learning},
  author={Al-Abbasi, Abubakr O and Ghosh, Arnob and Aggarwal, Vaneet},
  journal={IEEE Transactions on Intelligent Transportation Systems},
  volume={20},
  number={12},
  pages={4714--4727},
  year={2019},
  publisher={IEEE}
}

@article{zheng2018order,
  title={Order dispatch in price-aware ridesharing},
  author={Zheng, Libin and Chen, Lei and Ye, Jieping},
  journal={Proceedings of the VLDB Endowment},
  volume={11},
  number={8},
  pages={853--865},
  year={2018},
  publisher={VLDB Endowment}
}

@article{cap2018multi,
  title={Multi-objective analysis of ridesharing in automated mobility-on-demand},
  author={C{\'a}p, Michal and Alonso-Mora, Javier},
  journal={Proceedings Of RSS 2018: Robotics-Science And Systems XIV},
  year={2018}
}

@article{zardini2022co,
  title={Co-design to enable user-friendly tools to assess the impact of future mobility solutions},
  author={Zardini, Gioele and Lanzetti, Nicolas and Censi, Andrea and Frazzoli, Emilio and Pavone, Marco},
  journal={IEEE Transactions on Network Science and Engineering},
  volume={10},
  number={2},
  pages={827--844},
  year={2022},
  publisher={IEEE}
}

@techreport{iea2024globalEV,
  title        = {Global EV Outlook 2024: Moving Towards Increased Affordability},
  author       = {{International Energy Agency}},
  year         = {2024},
  institution  = {International Energy Agency},
  url          = {https://www.iea.org/reports/global-ev-outlook-2024},
  note         = {Accessed: [Date]}
}

@article{inci2015review,
  title={A review of the economics of parking},
  author={Inci, Eren},
  journal={Economics of Transportation},
  volume={4},
  number={1-2},
  pages={50--63},
  year={2015},
  publisher={Elsevier}
}

@article{becker2020impact,
  title={Impact of vehicle automation and electric propulsion on production costs for mobility services worldwide},
  author={Becker, Henrik and Becker, Felix and Abe, Ryosuke and Bekhor, Shlomo and Belgiawan, Prawira F and Compostella, Junia and Frazzoli, Emilio and Fulton, Lewis M and Bicudo, Davi Guggisberg and Gurumurthy, Krishna Murthy and others},
  journal={Transportation Research Part A: Policy and Practice},
  volume={138},
  pages={105--126},
  year={2020},
  publisher={Elsevier}
}

@article{sieber2020improved,
  title={Improved public transportation in rural areas with self-driving cars: A study on the operation of Swiss train lines},
  author={Sieber, Lukas and Ruch, Claudio and H{\"o}rl, Sebastian and Axhausen, Kay W and Frazzoli, Emilio},
  journal={Transportation research part A: policy and practice},
  volume={134},
  pages={35--51},
  year={2020},
  publisher={Elsevier}
}

@article{motzkin1956assignment,
  title={The assignment problem$^1$},
  author={Motzkin, TS},
  journal={Numerical analysis},
  number={6},
  pages={109},
  year={1956},
  publisher={American Mathematical Soc.}
}

@article{kuhn1955hungarian,
  title={The Hungarian method for the assignment problem},
  author={Kuhn, Harold W},
  journal={Naval research logistics quarterly},
  volume={2},
  number={1-2},
  pages={83--97},
  year={1955},
  publisher={Wiley Online Library}
}

@article{bertsekas1993parallel,
  title={Parallel asynchronous Hungarian methods for the assignment problem},
  author={Bertsekas, Dimitri P and Casta{\~n}on, David A},
  journal={ORSA Journal on Computing},
  volume={5},
  number={3},
  pages={261--274},
  year={1993},
  publisher={INFORMS}
}

@inproceedings{salazar2019congestion,
  title={A congestion-aware routing scheme for autonomous mobility-on-demand systems},
  author={Salazar, Mauro and Tsao, Matthew and Aguiar, Izabel and Schiffer, Maximilian and Pavone, Marco},
  booktitle={2019 18th European Control Conference (ECC)},
  pages={3040--3046},
  year={2019},
  organization={IEEE}
}

@article{braverman2019empty,
  title={Empty-car routing in ridesharing systems},
  author={Braverman, Anton and Dai, Jim G and Liu, Xin and Ying, Lei},
  journal={Operations Research},
  volume={67},
  number={5},
  pages={1437--1452},
  year={2019},
  publisher={INFORMS}
}

@article{liu2019dynamic,
  title={Dynamic shared autonomous taxi system considering on-time arrival reliability},
  author={Liu, Zhiguang and Miwa, Tomio and Zeng, Weiliang and Bell, Michael GH and Morikawa, Takayuki},
  journal={Transportation Research Part C: Emerging Technologies},
  volume={103},
  pages={281--297},
  year={2019},
  publisher={Elsevier}
}

@article{treleaven2013asymptotically,
  title={Asymptotically optimal algorithms for one-to-one pickup and delivery problems with applications to transportation systems},
  author={Treleaven, Kyle and Pavone, Marco and Frazzoli, Emilio},
  journal={IEEE Transactions on Automatic Control},
  volume={58},
  number={9},
  pages={2261--2276},
  year={2013},
  publisher={IEEE}
}

@article{ruch2020the+,
  title={The+ 1 method: model-free adaptive repositioning policies for robotic multi-agent systems},
  author={Ruch, Claudio and G{\"a}chter, Joel and Hakenberg, Jan and Frazzoli, Emilio},
  journal={IEEE Transactions on Network Science and Engineering},
  volume={7},
  number={4},
  pages={3171--3184},
  year={2020},
  publisher={IEEE}
}

@article{dalle2022learning,
  title={Learning with combinatorial optimization layers: a probabilistic approach},
  author={Dalle, Guillaume and Baty, L{\'e}o and Bouvier, Louis and Parmentier, Axel},
  journal={arXiv preprint arXiv:2207.13513},
  year={2022}
}

@inproceedings{schmidt2024learning,
  title={Learning to control autonomous fleets from observation via offline reinforcement learning},
  author={Schmidt, Carolin and Gammelli, Daniele and Pereira, Francisco Camara and Rodrigues, Filipe},
  booktitle={2024 European Control Conference (ECC)},
  pages={1399--1406},
  year={2024},
  organization={IEEE}
}

@inproceedings{iglesias2018data,
  title={Data-driven model predictive control of autonomous mobility-on-demand systems},
  author={Iglesias, Ramon and Rossi, Federico and Wang, Kevin and Hallac, David and Leskovec, Jure and Pavone, Marco},
  booktitle={2018 IEEE international conference on robotics and automation (ICRA)},
  pages={6019--6025},
  year={2018},
  organization={IEEE}
}

@inproceedings{tsao2018stochastic,
  title={Stochastic model predictive control for autonomous mobility on demand},
  author={Tsao, Matthew and Iglesias, Ramon and Pavone, Marco},
  booktitle={2018 21st International conference on intelligent transportation systems (ITSC)},
  pages={3941--3948},
  year={2018},
  organization={IEEE}
}

@article{he2023data,
  title={Data-driven distributionally robust electric vehicle balancing for autonomous mobility-on-demand systems under demand and supply uncertainties},
  author={He, Sihong and Zhang, Zhili and Han, Shuo and Pepin, Lynn and Wang, Guang and Zhang, Desheng and Stankovic, John A and Miao, Fei},
  journal={IEEE Transactions on Intelligent Transportation Systems},
  volume={24},
  number={5},
  pages={5199--5215},
  year={2023},
  publisher={IEEE}
}

@article{guo2022data,
  title={Data-driven vehicle rebalancing with predictive prescriptions in the ride-hailing system},
  author={Guo, Xiaotong and Wang, Qingyi and Zhao, Jinhua},
  journal={IEEE Open Journal of Intelligent Transportation Systems},
  volume={3},
  pages={251--266},
  year={2022},
  publisher={IEEE}
}

@article{carron2019scalable,
  title={Scalable model predictive control for autonomous mobility-on-demand systems},
  author={Carron, Andrea and Seccamonte, Francesco and Ruch, Claudio and Frazzoli, Emilio and Zeilinger, Melanie N},
  journal={IEEE Transactions on Control Systems Technology},
  volume={29},
  number={2},
  pages={635--644},
  year={2019},
  publisher={IEEE}
}

@article{sutton1992reinforcement,
  title={Reinforcement learning is direct adaptive optimal control},
  author={Sutton, Richard S and Barto, Andrew G and Williams, Ronald J},
  journal={IEEE control systems magazine},
  volume={12},
  number={2},
  pages={19--22},
  year={1992},
  publisher={IEEE}
}

@article{levine2020offline,
  title={Offline reinforcement learning: Tutorial, review, and perspectives on open problems},
  author={Levine, Sergey and Kumar, Aviral and Tucker, George and Fu, Justin},
  journal={arXiv preprint arXiv:2005.01643},
  year={2020}
}

@article{chouaki2025review,
  title={A review on reinforcement learning methods for mobility on demand systems},
  author={Chouaki, Tarek and H{\"o}rl, Sebastian and Puchinger, Jakob},
  journal={arXiv preprint arXiv:2501.02569},
  year={2025}
}

@inproceedings{gammelli2021graph,
  title={Graph neural network reinforcement learning for autonomous mobility-on-demand systems},
  author={Gammelli, Daniele and Yang, Kaidi and Harrison, James and Rodrigues, Filipe and Pereira, Francisco C and Pavone, Marco},
  booktitle={2021 60th IEEE Conference on Decision and Control (CDC)},
  pages={2996--3003},
  year={2021},
  organization={IEEE}
}

@inproceedings{gammelli2022graph,
  title={Graph meta-reinforcement learning for transferable autonomous mobility-on-demand},
  author={Gammelli, Daniele and Yang, Kaidi and Harrison, James and Rodrigues, Filipe and Pereira, Francisco and Pavone, Marco},
  booktitle={Proceedings of the 28th ACM SIGKDD Conference on Knowledge Discovery and Data Mining},
  pages={2913--2923},
  year={2022}
}

@article{liang2021integrated,
  title={An integrated reinforcement learning and centralized programming approach for online taxi dispatching},
  author={Liang, Enming and Wen, Kexin and Lam, William HK and Sumalee, Agachai and Zhong, Renxin},
  journal={IEEE Transactions on Neural Networks and Learning Systems},
  volume={33},
  number={9},
  pages={4742--4756},
  year={2021},
  publisher={IEEE}
}

@inproceedings{xu2018large,
  title={Large-scale order dispatch in on-demand ride-hailing platforms: A learning and planning approach},
  author={Xu, Zhe and Li, Zhixin and Guan, Qingwen and Zhang, Dingshui and Li, Qiang and Nan, Junxiao and Liu, Chunyang and Bian, Wei and Ye, Jieping},
  booktitle={Proceedings of the 24th ACM SIGKDD international conference on knowledge discovery \& data mining},
  pages={905--913},
  year={2018}
}

@article{baty2024combinatorial,
  title={Combinatorial optimization-enriched machine learning to solve the dynamic vehicle routing problem with time windows},
  author={Baty, L{\'e}o and Jungel, Kai and Klein, Patrick S and Parmentier, Axel and Schiffer, Maximilian},
  journal={Transportation Science},
  year={2024},
  publisher={INFORMS}
}

@article{maciejewski2016assignment,
  title={An assignment-based approach to efficient real-time city-scale taxi dispatching},
  author={Maciejewski, Michal and Bischoff, Joschka and Nagel, Kai},
  journal={IEEE Intelligent Systems},
  volume={31},
  number={1},
  pages={68--77},
  year={2016},
  publisher={IEEE}
}

@article{li2021real,
  title={A real-time dispatching strategy for shared automated electric vehicles with performance guarantees},
  author={Li, Li and Pantelidis, Theodoros and Chow, Joseph YJ and Jabari, Saif Eddin},
  journal={Transportation Research Part E: Logistics and Transportation Review},
  volume={152},
  pages={102392},
  year={2021},
  publisher={Elsevier}
}

@article{oh2020assessing,
  title={Assessing the impacts of automated mobility-on-demand through agent-based simulation: A study of Singapore},
  author={Oh, Simon and Seshadri, Ravi and Azevedo, Carlos Lima and Kumar, Nishant and Basak, Kakali and Ben-Akiva, Moshe},
  journal={Transportation Research Part A: Policy and Practice},
  volume={138},
  pages={367--388},
  year={2020},
  publisher={Elsevier}
}

@article{oh2021impacts,
  title={Impacts of Automated Mobility-on-Demand on traffic dynamics, energy and emissions: A case study of Singapore},
  author={Oh, Simon and Lentzakis, Antonis F and Seshadri, Ravi and Ben-Akiva, Moshe},
  journal={Simulation Modelling Practice and Theory},
  volume={110},
  pages={102327},
  year={2021},
  publisher={Elsevier}
}

@article{wardrop1952road,
  title={Road paper. some theoretical aspects of road traffic research.},
  author={Wardrop, John Glen},
  journal={Proceedings of the institution of civil engineers},
  volume={1},
  number={3},
  pages={325--362},
  year={1952},
  publisher={Thomas Telford-ICE Virtual Library}
}

@book{us1964traffic,
  title={Traffic assignment manual for application with a large, high speed computer},
  author={{US Bureau of Public Roads}},
  year={1964},
  publisher={US Department of Commerce}
}

@article{solovey2019scalable,
  title={Scalable and congestion-aware routing for autonomous mobility-on-demand via frank-wolfe optimization},
  author={Solovey, Kiril and Salazar, Mauro and Pavone, Marco},
  journal={arXiv preprint arXiv:1903.03697},
  year={2019}
}

@article{daganzo2008analytical,
  title={An analytical approximation for the macroscopic fundamental diagram of urban traffic},
  author={Daganzo, Carlos F and Geroliminis, Nikolas},
  journal={Transportation Research Part B: Methodological},
  volume={42},
  number={9},
  pages={771--781},
  year={2008},
  publisher={Elsevier}
}

@article{daganzo1994cell,
  title={The cell transmission model: A dynamic representation of highway traffic consistent with the hydrodynamic theory},
  author={Daganzo, Carlos F},
  journal={Transportation research part B: methodological},
  volume={28},
  number={4},
  pages={269--287},
  year={1994},
  publisher={Elsevier}
}

@inproceedings{yperman2005link,
  title={The link transmission model: An efficient implementation of the kinematic wave theory in traffic networks},
  author={Yperman, Isaak and Logghe, Steven and Immers, Ben},
  booktitle={Proceedings of the 10th EWGT Meeting},
  volume={24},
  pages={122--127},
  year={2005},
  organization={Citeseer}
}

@article{liu2019framework,
  title={A framework to integrate mode choice in the design of mobility-on-demand systems},
  author={Liu, Yang and Bansal, Prateek and Daziano, Ricardo and Samaranayake, Samitha},
  journal={Transportation Research Part C: Emerging Technologies},
  volume={105},
  pages={648--665},
  year={2019},
  publisher={Elsevier}
}

@article{james2019online,
  title={Online vehicle routing with neural combinatorial optimization and deep reinforcement learning},
  author={James, JQ and Yu, Wen and Gu, Jiatao},
  journal={IEEE Transactions on Intelligent Transportation Systems},
  volume={20},
  number={10},
  pages={3806--3817},
  year={2019},
  publisher={IEEE}
}

@misc{nyc_tlc_trip_data,
  author       = "{New York City Taxi and Limousine Commission}",
  title        = "{TLC Trip Record Data}",
  year         = 2025,
  howpublished = "\url{https://www.nyc.gov/taxi}",
  note         = "Accessed: 2025-01-15"
}

@misc{illinoisdatabankIDB-9610843,
doi = {10.13012/J8PN93H8},
url = {https://doi.org/10.13012/J8PN93H8},
author = {Donovan, Brian and Work, Dan},
publisher = {University of Illinois Urbana-Champaign},
title = {New York City Taxi Trip Data (2010-2013)},
year = {2016}
}

@article{donovan2015using,
  title={Using coarse GPS data to quantify city-scale transportation system resilience to extreme events},
  author={Donovan, Brian and Work, Daniel B},
  journal={arXiv preprint arXiv:1507.06011},
  year={2015}
}

@misc{donovan_published_code,
  author       = {Brian Donovan},
  title        = {Published code},
  howpublished = {\url{https://github.com/Lab-Work/gpsresilience}},
  note         = {Accessed: 2025-01-14}
}

@inproceedings{sadeghi2022reinforcement,
  title={Reinforcement learning in the wild: Scalable RL dispatching algorithm deployed in ridehailing marketplace},
  author={Sadeghi Eshkevari, Soheil and Tang, Xiaocheng and Qin, Zhiwei and Mei, Jinhan and Zhang, Cheng and Meng, Qianying and Xu, Jia},
  booktitle={Proceedings of the 28th ACM SIGKDD Conference on Knowledge Discovery and Data Mining},
  pages={3838--3848},
  year={2022}
}

@misc{c7j010-22,
doi = {10.15783/C7J010},
url = {https://dx.doi.org/10.15783/C7J010},
author = {Piorkowski, Michal and Sarafijanovic-Djukic, Natasa and Grossglauser, Matthias },
publisher = {IEEE Dataport},
title = {CRAWDAD epfl/mobility},
year = {2022} }

@misc{transportation_networks,
  author       = {Transportation Networks for Research Core Team},
  title        = {Transportation Networks for Research},
  howpublished = {\url{https://github.com/bstabler/TransportationNetworks}},
  note         = {Accessed: January 14, 2025}
}

@book{w2016multi,
  title={The multi-agent transport simulation MATSim},
  author={W Axhausen, Kay and Horni, Andreas and Nagel, Kai},
  year={2016},
  publisher={Ubiquity Press}
}

@inproceedings{adnan2016simmobility,
  title={Simmobility: A multi-scale integrated agent-based simulation platform},
  author={Adnan, Muhammad and Pereira, Francisco C and Azevedo, Carlos Miguel Lima and Basak, Kakali and Lovric, Milan and Raveau, Sebasti{\'a}n and Zhu, Yi and Ferreira, Joseph and Zegras, Christopher and Ben-Akiva, Moshe},
  booktitle={95th Annual Meeting of the Transportation Research Board Forthcoming in Transportation Research Record},
  volume={2},
  year={2016},
  organization={The National Academies of Sciences, Engineering, and Medicine Washington, DC}
}

@inproceedings{ruch2018amodeus,
  title={Amodeus, a simulation-based testbed for autonomous mobility-on-demand systems},
  author={Ruch, Claudio and H{\"o}rl, Sebastian and Frazzoli, Emilio},
  booktitle={2018 21st international conference on intelligent transportation systems (ITSC)},
  pages={3639--3644},
  year={2018},
  organization={IEEE}
}

@inproceedings{lopez2018microscopic,
  title={Microscopic traffic simulation using sumo},
  author={Lopez, Pablo Alvarez and Behrisch, Michael and Bieker-Walz, Laura and Erdmann, Jakob and Fl{\"o}tter{\"o}d, Yun-Pang and Hilbrich, Robert and L{\"u}cken, Leonhard and Rummel, Johannes and Wagner, Peter and Wie{\ss}ner, Evamarie},
  booktitle={2018 21st international conference on intelligent transportation systems (ITSC)},
  pages={2575--2582},
  year={2018},
  organization={IEEE}
}

@inproceedings{marczuk2016simulation,
  title={Simulation framework for rebalancing of autonomous mobility on demand systems},
  author={Marczuk, Katarzyna A and Soh, Harold SH and Azevedo, Carlos ML and Lee, Der-Horng and Frazzoli, Emilio},
  booktitle={MATEC Web of Conferences},
  volume={81},
  pages={01005},
  year={2016},
  organization={EDP Sciences}
}

@article{nguyen2023examining,
  title={Examining the effects of Automated Mobility-on-Demand services on public transport systems using an agent-based simulation approach},
  author={Nguyen-Phuoc, Duy Q and Zhou, Meng and Chua, Ming Hong and Alho, Andr{\'e} Romano and Oh, Simon and Seshadri, Ravi and Le, Diem-Trinh},
  journal={Transportation Research Part A: Policy and Practice},
  volume={169},
  pages={103583},
  year={2023},
  publisher={Elsevier}
}

@article{nahmias2021evaluating,
  title={Evaluating the impacts of shared automated mobility on-demand services: An activity-based accessibility approach},
  author={Nahmias-Biran, Bat-hen and Oke, Jimi B and Kumar, Nishant and Lima Azevedo, Carlos and Ben-Akiva, Moshe},
  journal={Transportation},
  volume={48},
  pages={1613--1638},
  year={2021},
  publisher={Springer}
}

@article{zhou2021simulating,
  title={Simulating impacts of Automated Mobility-on-Demand on accessibility and residential relocation},
  author={Zhou, Meng and Le, Diem-Trinh and Nguyen-Phuoc, Duy Quy and Zegras, P Christopher and Ferreira Jr, Joseph},
  journal={Cities},
  volume={118},
  pages={103345},
  year={2021},
  publisher={Elsevier}
}

@misc{amodeus_repository,
  author       = {AMoDeus Science Team},
  title        = {AMoDeus: A Framework for Autonomous Mobility-on-Demand Systems},
  howpublished = {\url{https://github.com/amodeus-science/amodeus/tree/master?tab=readme-ov-file}},
  note         = {Accessed: January 14, 2025}
}

@inproceedings{zardini2023strategic,
  title={Strategic interactions in multi-modal mobility systems: A game-theoretic perspective},
  author={Zardini, Gioele and Lanzetti, Nicolas and Belgioioso, Giuseppe and Hartnik, Christian and Bolognani, Saverio and D{\"o}rfler, Florian and Frazzoli, Emilio},
  booktitle={2023 IEEE 26th International Conference on Intelligent Transportation Systems (ITSC)},
  pages={5452--5459},
  year={2023},
  organization={IEEE}
}

@INPROCEEDINGS{9564501,
  author={Zardini, Gioele and Lanzetti, Nicolas and Guerrini, Laura and Frazzoli, Emilio and Dörfler, Florian},
  booktitle={2021 IEEE International Intelligent Transportation Systems Conference (ITSC)}, 
  title={Game Theory to Study Interactions between Mobility Stakeholders}, 
  year={2021},
  volume={},
  number={},
  pages={2054-2061},
  doi={10.1109/ITSC48978.2021.9564501}}

@article{oke2020evaluating,
  title={Evaluating the systemic effects of automated mobility-on-demand services via large-scale agent-based simulation of auto-dependent prototype cities},
  author={Oke, Jimi B and Akkinepally, Arun Prakash and Chen, Siyu and Xie, Yifei and Aboutaleb, Youssef M and Azevedo, Carlos Lima and Zegras, P Christopher and Ferreira, Joseph and Ben-Akiva, Moshe},
  journal={Transportation Research Part A: Policy and Practice},
  volume={140},
  pages={98--126},
  year={2020},
  publisher={Elsevier}
}

@misc{transportation_decarbonization,
  author       = {J-PAL (Abdul Latif Jameel Poverty Action Lab)},
  title        = {Transportation Decarbonization White Paper},
  howpublished = {\url{https://www.povertyactionlab.org/publication/transportation-decarbonization-white-paper}},
  note         = {Accessed: January 14, 2025}
}

@inproceedings{zardini2020co,
  title={On the co-design of AV-enabled mobility systems},
  author={Zardini, Gioele and Lanzetti, Nicolas and Salazar, Mauro and Censi, Andrea and Frazzoli, Emilio and Pavone, Marco},
  booktitle={2020 IEEE 23rd International Conference on Intelligent Transportation Systems (ITSC)},
  pages={1--8},
  year={2020},
  organization={IEEE}
}

@inproceedings{zardini2021co,
  title={Co-design of embodied intelligence: A structured approach},
  author={Zardini, Gioele and Milojevic, Dejan and Censi, Andrea and Frazzoli, Emilio},
  booktitle={2021 IEEE/RSJ International Conference on Intelligent Robots and Systems (IROS)},
  pages={7536--7543},
  year={2021},
  organization={IEEE}
}

@inproceedings{zardini2021co_hardware,
  title={Co-design of autonomous systems: From hardware selection to control synthesis},
  author={Zardini, Gioele and Censi, Andrea and Frazzoli, Emilio},
  booktitle={2021 European Control Conference (ECC)},
  pages={682--689},
  year={2021},
  organization={IEEE}
}

@article{kamel2019exploring,
  title={Exploring the impact of user preferences on shared autonomous vehicle modal split: A multi-agent simulation approach},
  author={Kamel, Joseph and Vosooghi, Reza and Puchinger, Jakob and Ksontini, Feirouz and Sirin, G{\"o}knur},
  journal={Transportation Research Procedia},
  volume={37},
  pages={115--122},
  year={2019},
  publisher={Elsevier}
}

@article{steck2018autonomous,
  title={How autonomous driving may affect the value of travel time savings for commuting},
  author={Steck, Felix and Kolarova, Viktoriya and Bahamonde-Birke, Francisco and Trommer, Stefan and Lenz, Barbara},
  journal={Transportation research record},
  volume={2672},
  number={46},
  pages={11--20},
  year={2018},
  publisher={SAGE Publications Sage CA: Los Angeles, CA}
}

@article{liu2017tracking,
  title={Tracking a system of shared autonomous vehicles across the Austin, Texas network using agent-based simulation},
  author={Liu, Jun and Kockelman, Kara M and Boesch, Patrick M and Ciari, Francesco},
  journal={Transportation},
  volume={44},
  pages={1261--1278},
  year={2017},
  publisher={Springer}
}

@inproceedings{kagho2024framework,
  title={Framework for analyzing equity-concerns related to mobility on-demand},
  author={Kagho, Grace Orowo and Murthy Gurumurthy, Krishna and Verbas, {\"O}mer and Auld, Joshua},
  booktitle={103rd Annual Meeting of the Transportation Research Board (TRB 2024)},
  year={2024},
  organization={IVT ETH Zurich}
}

@book{litman2017evaluating,
  title={Evaluating transportation equity},
  author={Litman, Todd},
  year={2017},
  publisher={Victoria Transport Policy Institute Victoria, BC, Canada}
}

@article{bhuyan2019gis,
  title={GIS-based equity gap analysis: Case study of Baltimore bike share program},
  author={Bhuyan, Istiak A and Chavis, Celeste and Nickkar, Amirreza and Barnes, Philip},
  journal={Urban Science},
  volume={3},
  number={2},
  pages={42},
  year={2019},
  publisher={MDPI}
}

@techreport{schrank20112011,
  title={2011 urban mobility report},
  author={Schrank, David and Lomax, Tim and Eisele, Bill and others},
  year={2011},
  institution={Texas Transportation Institute}
}

@article{munafo2017manifesto,
  title={A manifesto for reproducible science},
  author={Munaf{\`o}, Marcus R and Nosek, Brian A and Bishop, Dorothy VM and Button, Katherine S and Chambers, Christopher D and Percie du Sert, Nathalie and Simonsohn, Uri and Wagenmakers, Eric-Jan and Ware, Jennifer J and Ioannidis, John},
  journal={Nature human behaviour},
  volume={1},
  number={1},
  pages={1--9},
  year={2017},
  publisher={Nature Publishing Group}
}

@article{peng2011reproducible,
  title={Reproducible research in computational science},
  author={Peng, Roger D},
  journal={Science},
  volume={334},
  number={6060},
  pages={1226--1227},
  year={2011},
  publisher={American Association for the Advancement of Science}
}

@inproceedings{leitner2017acrv,
  title={The ACRV picking benchmark: A robotic shelf picking benchmark to foster reproducible research},
  author={Leitner, J{\"u}rgen and Tow, Adam W and S{\"u}nderhauf, Niko and Dean, Jake E and Durham, Joseph W and Cooper, Matthew and Eich, Markus and Lehnert, Christopher and Mangels, Ruben and McCool, Christopher and others},
  booktitle={2017 IEEE international conference on robotics and automation (ICRA)},
  pages={4705--4712},
  year={2017},
  organization={IEEE}
}

@article{baines2024need,
  title={The need for reproducible research in soft robotics},
  author={Baines, Robert and Shah, Dylan and Marvel, Jeremy and Case, Jennifer and Spielberg, Andrew},
  journal={Nature Machine Intelligence},
  volume={6},
  number={7},
  pages={740--741},
  year={2024},
  publisher={Nature Publishing Group UK London}
}

@inproceedings{chen2013electric,
  title={The electric vehicle charging station location problem: a parking-based assignment method for Seattle},
  author={Chen, T Donna and Kockelman, Kara M and Khan, Moby and others},
  booktitle={Transportation Research Board 92nd Annual Meeting},
  volume={340},
  pages={13--1254},
  year={2013}
}

@inproceedings{jia2012optimal,
  title={Optimal siting and sizing of electric vehicle charging stations},
  author={Jia, Long and Hu, Zechun and Song, Yonghua and Luo, Zhuowei},
  booktitle={2012 IEEE International Electric Vehicle Conference},
  pages={1--6},
  year={2012},
  organization={IEEE}
}

@article{zhang2016pev,
  title={PEV fast-charging station siting and sizing on coupled transportation and power networks},
  author={Zhang, Hongcai and Moura, Scott J and Hu, Zechun and Song, Yonghua},
  journal={IEEE Transactions on Smart Grid},
  volume={9},
  number={4},
  pages={2595--2605},
  year={2016},
  publisher={IEEE}
}

@inproceedings{luke2021joint,
  title={Joint optimization of autonomous electric vehicle fleet operations and charging station siting},
  author={Luke, Justin and Salazar, Mauro and Rajagopal, Ram and Pavone, Marco},
  booktitle={2021 IEEE International Intelligent Transportation Systems Conference (ITSC)},
  pages={3340--3347},
  year={2021},
  organization={IEEE}
}

@article{zardini2022analysis,
  title={Analysis and control of autonomous mobility-on-demand systems},
  author={Zardini, Gioele and Lanzetti, Nicolas and Pavone, Marco and Frazzoli, Emilio},
  journal={Annual Review of Control, Robotics, and Autonomous Systems},
  volume={5},
  number={1},
  pages={633--658},
  year={2022},
  publisher={Annual Reviews}
}

@ARTICLE{9924240,
  author={Tan, Wenrui and Sun, Yimeng and Ding, Zhaohao and Lee, Wei-Jen},
  journal={IEEE Open Access Journal of Power and Energy}, 
  title={Fleet Management and Charging Scheduling for Shared Mobility-on-Demand System: A Systematic Review}, 
  year={2022},
  volume={9},
  number={},
  pages={425-436},
  doi={10.1109/OAJPE.2022.3215865}}

@inproceedings{zhang2015queueing,
  title={A queueing network approach to the analysis and control of mobility-on-demand systems},
  author={Zhang, Rick and Pavone, Marco},
  booktitle={2015 American Control Conference (ACC)},
  pages={4702--4709},
  year={2015},
  organization={IEEE}
}

@article{semmelrock2023reproducibility,
  title={Reproducibility in machine learning-driven research},
  author={Semmelrock, Harald and Kopeinik, Simone and Theiler, Dieter and Ross-Hellauer, Tony and Kowald, Dominik},
  journal={arXiv preprint arXiv:2307.10320},
  year={2023}
}

@article{riehl2025revisiting,
  title={Revisiting reproducibility in transportation simulation studies},
  author={Riehl, Kevin and Kouvelas, Anastasios and Makridis, Michail A},
  journal={European Transport Research Review},
  volume={17},
  number={1},
  pages={22},
  year={2025},
  publisher={Springer}
}

@inproceedings{imagenet_cvpr09,
  author    = {Jia Deng and Wei Dong and Richard Socher and Li-Jia Li and Kai Li and Li Fei-Fei},
  title     = {ImageNet: A Large-Scale Hierarchical Image Database},
  booktitle = {2009 IEEE Conference on Computer Vision and Pattern Recognition (CVPR)},
  year      = {2009},
  pages     = {248--255},
  doi       = {10.1109/CVPR.2009.5206848}
}

@inproceedings{geiger2012we,
  title     = {Are we ready for Autonomous Driving? The KITTI Vision Benchmark Suite},
  author    = {Andreas Geiger and Philip Lenz and Raquel Urtasun},
  booktitle = {2012 IEEE Conference on Computer Vision and Pattern Recognition (CVPR)},
  year      = {2012},
  pages     = {3354--3361},
  doi       = {10.1109/CVPR.2012.6248074}
}

@inproceedings{ahmed2022managing,
  title={Managing randomness to enable reproducible machine learning},
  author={Ahmed, Hana and Lofstead, Jay},
  booktitle={Proceedings of the 5th International Workshop on practical reproducible evaluation of computer systems},
  pages={15--20},
  year={2022}
}

@misc{stanfordasl_rl4amod_2024,
  author       = {Zardini, Gioele and Gammelli, Daniele and Tresca, Luigi and Schmidt, Carolin and Harrison, James and Rodrigues, Filipe and Schiffer, Maximilian and Pavone, Marco},
  title        = {Data-driven Methods for Network-level Coordination of Autonomous Mobility-on-Demand Systems Across Scales},
  year         = {2024},
  howpublished = {\url{https://github.com/StanfordASL/RL4AMOD}},
  note         = {Accessed: 2025-04-17}
}

@article{tresca2025robo,
  title={Robo-taxi Fleet Coordination at Scale via Reinforcement Learning},
  author={Tresca, Luigi and Schmidt, Carolin and Harrison, James and Rodrigues, Filipe and Zardini, Gioele and Gammelli, Daniele and Pavone, Marco},
  journal={arXiv preprint arXiv:2504.06125},
  year={2025}
}

@article{bellman1966dynamic,
  title={Dynamic programming},
  author={Bellman, Richard},
  journal={science},
  volume={153},
  number={3731},
  pages={34--37},
  year={1966},
  publisher={American Association for the Advancement of Science}
}

@article{gammelli2020estimating,
  title={Estimating latent demand of shared mobility through censored Gaussian processes},
  author={Gammelli, Daniele and Peled, Inon and Rodrigues, Filipe and Pacino, Dario and Kurtaran, Haci A and Pereira, Francisco C},
  journal={Transportation Research Part C: Emerging Technologies},
  volume={120},
  pages={102775},
  year={2020},
  publisher={Elsevier}
}

@article{hussein2017imitation,
  title={Imitation learning: A survey of learning methods},
  author={Hussein, Ahmed and Gaber, Mohamed Medhat and Elyan, Eyad and Jayne, Chrisina},
  journal={ACM Computing Surveys (CSUR)},
  volume={50},
  number={2},
  pages={1--35},
  year={2017},
  publisher={ACM New York, NY, USA}
}

@article{engelhardt2022fleetpy,
  title={Fleetpy: A modular open-source simulation tool for mobility on-demand services},
  author={Engelhardt, Roman and Dandl, Florian and Syed, Arslan-Ali and Zhang, Yunfei and Fehn, Fabian and Wolf, Fynn and Bogenberger, Klaus},
  journal={arXiv preprint arXiv:2207.14246},
  year={2022}
}

@book{kounev2020systems,
  title={Systems Benchmarking},
  author={Kounev, Samuel and Lange, Klaus-Dieter and Von Kistowski, Joakim},
  volume={1},
  year={2020},
  publisher={Springer}
}

@inproceedings{schmidt2024offline,
  title={Offline Hierarchical Reinforcement Learning via Inverse Optimization},
  author={Schmidt, Carolin and Gammelli, Daniele and Harrison, James and Pavone, Marco and Rodrigues, Filipe},
  booktitle={International Conference on Learning Representations (ICLR)},
  year={2025}
}

@article{fivsar2024reproducibility,
  title={Reproducibility in management science},
  author={Fi{\v{s}}ar, Milo{\v{s}} and Greiner, Ben and Huber, Christoph and Katok, Elena and Ozkes, Ali I and Management Science Reproducibility Collaboration},
  journal={Management Science},
  volume={70},
  number={3},
  pages={1343--1356},
  year={2024},
  publisher={INFORMS}
}

@article{molter2024public,
  title={Public transport across models and scales: A case study of the Munich network},
  author={M{\"o}lter, Jan and Ji, Joanna and Lienkamp, Benedikt and Zhang, Qin and Moreno, Ana T and Schiffer, Maximilian and Moeckel, Rolf and Kuehn, Christian},
  journal={PNAS nexus},
  volume={3},
  number={11},
  pages={pgae489},
  year={2024},
  publisher={Oxford University Press US}
}

@article{lienkamp2024column,
  title={Column generation for solving large scale multi-commodity flow problems for passenger transportation},
  author={Lienkamp, Benedikt and Schiffer, Maximilian},
  journal={European Journal of Operational Research},
  volume={314},
  number={2},
  pages={703--717},
  year={2024},
  publisher={Elsevier}
}

@article{jalota2023online,
  title={Online routing over parallel networks: Deterministic limits and data-driven enhancements},
  author={Jalota, Devansh and Paccagnan, Dario and Schiffer, Maximilian and Pavone, Marco},
  journal={INFORMS Journal on Computing},
  volume={35},
  number={3},
  pages={560--577},
  year={2023},
  publisher={INFORMS}
}

@article{coppola2024staggered,
  title={Staggered Routing in Autonomous Mobility-on-Demand Systems},
  author={Coppola, Antonio and Hiermann, Gerhard and Paccagnan, Dario and Schiffer, Maximilian},
  journal={arXiv preprint arXiv:2405.01410},
  year={2024}
}

@article{klein2023electric,
  title={Electric vehicle charge scheduling with flexible service operations},
  author={Klein, Patrick S and Schiffer, Maximilian},
  journal={Transportation Science},
  volume={57},
  number={6},
  pages={1605--1626},
  year={2023},
  publisher={INFORMS}
}

@misc{didiglobal2025,
  author       = {{Didi Global Inc.}},
  title        = {Didi Global Official Website},
  howpublished = {\url{https://web.didiglobal.com/}},
  note         = {Accessed: 2025-05-22},
  year         = {2025}
}

@article{goodman2016does,
  title={What does research reproducibility mean?},
  author={Goodman, Steven N and Fanelli, Daniele and Ioannidis, John PA},
  journal={Science translational medicine},
  volume={8},
  number={341},
  pages={341ps12--341ps12},
  year={2016},
  publisher={American Association for the Advancement of Science}
}

@book{national2019reproducibility,
  title={Reproducibility and replicability in science},
  author={National Academies of Sciences and Medicine and Policy and Global Affairs and Board on Research Data and Information and Division on Engineering and Physical Sciences and Committee on Applied and Theoretical Statistics and others},
  year={2019},
  publisher={National Academies Press}
}

@article{beiranvand2017best,
  title={Best practices for comparing optimization algorithms},
  author={Beiranvand, Vahid and Hare, Warren and Lucet, Yves},
  journal={Optimization and Engineering},
  volume={18},
  pages={815--848},
  year={2017},
  publisher={Springer}
}

@article{hoppe2025structured,
  title={Structured Reinforcement Learning for Combinatorial Decision-Making},
  author={Hoppe, Heiko and Baty, L{\'e}o and Bouvier, Louis and Parmentier, Axel and Schiffer, Maximilian},
  journal={arXiv preprint arXiv:2505.19053},
  year={2025}
}
